\pdfoutput=1

\documentclass[11pt]{article}

\usepackage[]{ACL2023}

\usepackage{times}
\usepackage{latexsym}
\usepackage{amsmath}
\usepackage{adjustbox}
\usepackage{verbatim}
\usepackage[T1]{fontenc}
\usepackage{xcolor}

\usepackage[utf8]{inputenc}

\usepackage{microtype}
\usepackage{booktabs}
\usepackage{multirow}
\usepackage{inconsolata}

\title{Revisiting Relation Extraction in the era of Large Language Models}

\author{\textbf{Somin Wadhwa}\quad\quad \textbf{Silvio Amir}\quad\quad \textbf{Byron C. Wallace} \\ 
Northeastern University  \\ 
\texttt{\{wadhwa.s, s.amir, b.wallace\}@northeastern.edu} 
}

\begin{document}
\maketitle
\begin{abstract}

Relation extraction (RE) is the core NLP task of inferring semantic relationships between entities from text. 
Standard supervised RE techniques entail training modules to tag tokens comprising entity spans and then predict the relationship between them. 
Recent work has instead treated the problem as a \emph{sequence-to-sequence} task, linearizing relations between entities as target strings to be generated conditioned on the input. 
Here we push the limits of this approach, using larger language models (GPT-3 and Flan-T5 large) than considered in prior work and evaluating their performance on standard RE tasks under varying levels of supervision.
We address issues inherent to evaluating generative approaches to RE by doing human evaluations, 
in lieu of relying on exact matching.
Under this refined evaluation, we find that: (1) \emph{Few-shot} prompting with GPT-3 achieves near SOTA performance, i.e., roughly equivalent to existing \emph{fully supervised} models; (2) Flan-T5 is not as capable in the few-shot setting, but supervising and fine-tuning it with Chain-of-Thought (CoT) style explanations (generated via GPT-3) yields SOTA results. 
We release this model as a new baseline for RE tasks\footnote{\url{https://sominw.com/ACL23LLMs}}.

\end{abstract}

\section{Introduction}

\begin{figure}
    \centering
\includegraphics[scale=0.060]{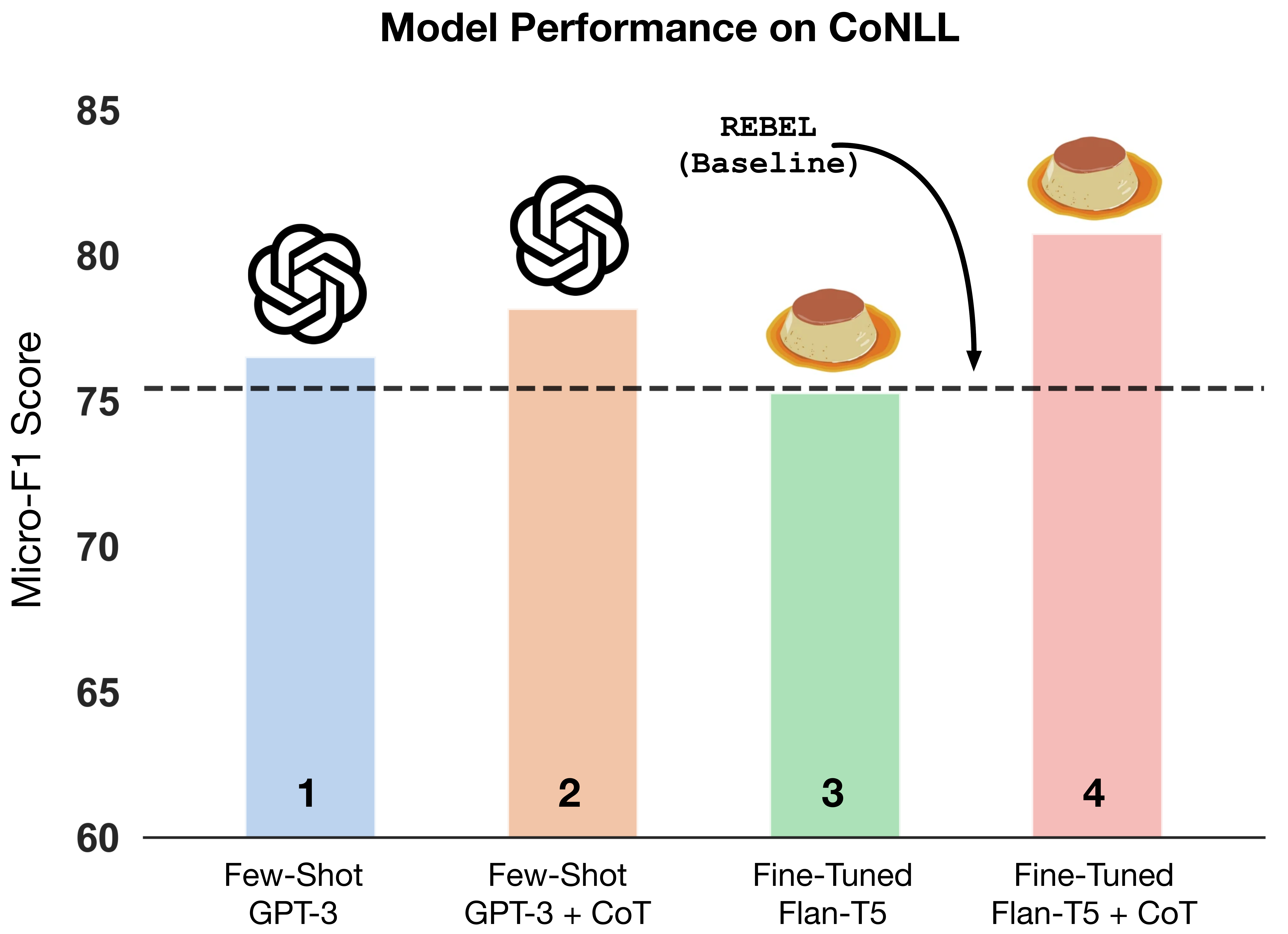}
\caption{RE performance of LLMs on the CoNLL dataset. {\bf 1} \emph{Few-shot} GPT-3 slightly outperforms the existing \emph{fully supervised} SOTA method (\citealt{huguet-cabot-navigli-2021-rebel-relation}; dotted horizontal line). {\bf 2} Eliciting CoT reasoning from GPT-3 further improves few-shot performance. {\bf 3} Fine-tuning Flan-T5 (large) is competitive with, but no better than, existing supervised methods, but {\bf 4} supervising Flan-T5 with CoT reasoning elicited from GPT-3 substantially outperforms all other models.
} 
\label{fig:conllres}
\end{figure}

\emph{Relation extraction} (RE) is the task of identifying entities and their semantic relationships from texts. 
Standard supervised approaches \cite{Eberts2019SpanbasedJE} to RE learn to tag entity spans and then classify relationships (if any) between these. 
More recent work has shown that conditional language models can capably perform this task---achieving SOTA or near-SOTA results---when trained to output linearized strings encoding entity pairs and their relations \cite{paolini2021structured, lu-etal-2022-unified, huguet-cabot-navigli-2021-rebel-relation}.
However, to date such work has considered only moderately sized pre-trained models for RE such as BART \cite{paolini2021structured, huguet-cabot-navigli-2021-rebel-relation}.

In this work we investigate the use of very large language models----including GPT-3 \cite{Brown2020LanguageMA}---for end-to-end relation extraction via generation. 
Our contributions are as follows. 

    \vspace{0.35em}
\noindent 1. We show that few-shot learning with GPT-3 yields near SOTA performance on standard RE datasets, outperforming fully supervised models. 

\vspace{0.35em}
\noindent 2. We find that Flan-T5 (large; \citealt{https://doi.org/10.48550/arxiv.2210.11416}) is not as capable, even when fine-tuned. 
But we then propose an approach to training Flan-T5 with \emph{Chain-of-Thought} (CoT) style ``explanations'' (generated automatically by GPT-3) that support relation inferences; this achieves SOTA results. 

  
 \vspace{0.35em}
 \noindent 3. Evaluating the performance of \emph{generative} models for RE is non-trivial because one cannot rely on exact matches to targets. We address this by collecting a small amount of annotations scoring generated outputs against targets. We use these annotations to quantify the problem, identify erroneous gold references and accurately evaluate our models.
 
 
\vspace{0.35em}
\noindent Our results indicate that, in general, {\bf LLMs should be the default approach to RE}, especially given that one can train Flan-T5---which is dramatically smaller than GPT-3, and publicly available---to achieve SOTA performance (Figure \ref{fig:conllres}).


\section{RE via Text Generation}

We treat RE as a conditional text generation task. 
Concretely, for a dataset of size $N$, we model the probability of generating a \emph{linearized} string $y$ of a relation triplet (\texttt{entity$\_$1, relation$\_$type, entity$\_2$}) conditioned on a context string $\mathcal{C}$. 
Specifically, $\mathcal{C}$ includes a chain of $n$ linearized examples $(x_i, y_i)$, with $n << N$.
Formally: 
\begin{equation*}
    p_{\text{LM}}(y | \mathcal{C}, x) = \prod_{t=1}^{T}p(y_t | \mathcal{C}, x, y_{<t})
\end{equation*}
We provide examples of context strings in the Appendix. 
We conduct experiments over four standard RE datasets comprising varying numbers of entities and relation types, namely ADE \cite{Gurulingappa2012DevelopmentOA}, CoNLL \cite{roth-yih-2004-linear}, NYT \cite{Riedel2010ModelingRA}, and DocRED (\citealt{yao-etal-2019-docred}); details in Table \ref{tab:datastats} and Appendix \ref{appendix:datasets}.

Following \citet{huguet-cabot-navigli-2021-rebel-relation}, we linearize our target relation triplets. 
However, we adopt a much simpler scheme than prior work: We linearize inputs with a single relation type (e.g. ADE) as a list of tuples:
\begin{center}
    \texttt{[(drug, effect), ... ,(drug, effect)]}
\end{center}

For inputs with multiple relation types (as in CoNLL04 and NYT), we form \textit{triplets} comprising a \texttt{subject}, \texttt{relation}, and \texttt{object} (along with their corresponding types), in the order of appearance of the subject entity: 

\begin{center}\texttt{[(entity$\_$1:entity$\_$1$\_$type, relation$\_$type, entity$\_$2:entity$\_$2$\_$type),..]}
\end{center}

A training instance is then a pair of input text and a linearized target string: 

\begin{flushleft}
\texttt{\textbf{Input }Bill Nelson, NASA administrator announced the mars mission today.}\\
\texttt{\textbf{Target }[(Bill Nelson:Per, Work$\_$For, NASA:Org)]}
\end{flushleft}

\begin{table}[]
\small
\centering
\begin{tabular}{@{}lrrrrr@{}}
\toprule
        & \multicolumn{1}{c}{\multirow{2}{*}{\textbf{\begin{tabular}[c]{@{}c@{}}Entity\\ Types\end{tabular}}}} & \multicolumn{1}{c}{\multirow{2}{*}{\textbf{\begin{tabular}[c]{@{}c@{}}Relation\\ Types\end{tabular}}}} & \multicolumn{3}{c}{$\#$ of relation triplets}                                                      \\
        & \multicolumn{1}{c}{}                                                                                   & \multicolumn{1}{c}{}                                                                                    & \multicolumn{1}{r}{\textbf{Train}} & \multicolumn{1}{r}{\textbf{Val}} & \multicolumn{1}{r}{\textbf{Test}} \\ \cmidrule(l){2-6} 
ADE     & 2                                                                                                      & 1                                                                                                       & 4,272                              & --                                      & --                                \\
CoNLL04 & 4                                                                                                      & 5                                                                                                       & 922                                & 231                                     & 288                               \\
NYT     & 4                                                                                                      & 24                                                                                                      & 56,196                             & 5,000                                   & 5,000                             \\
DocRED  & 6                                                                                                      & 96                                                                                                      & 3,008                              & 300                                     & 700                               \\ \bottomrule
\end{tabular}
\caption{Dataset statistics. Train, validation and test indicate the number of relation triplets in each dataset.} 
\label{tab:datastats}
\end{table}

\paragraph{Challenges inherent to evaluating generative large language models for RE} The expressivity of language models coupled with the open-endedness of RE makes evaluation difficult.
This has led to inconsistent approaches to evaluation \cite{taille-etal-2020-lets}.
Past work, especially that pre-dating LLMs for the task, has tended to perform ``strict'' evaluation, requiring exact matches between generated linearized relation tuples and references. 
This may be appropriate when is evaluating smaller conditional generation models (such as BART) for RE, which have been \emph{fine-tuned} on large training sets, because after training such models consistently generate standardized outputs.
By contrast, however, models like GPT-3 (or other large language models capable of zero- or few-shot application) can produce a wide variety of output formats which convey similar content.

For example, given an input from ADE and prompted to \emph{list all drugs and associated adverse events}, a large language model might yield \emph{Aspirin: stomach pain, chest pain}. Or it may instead output: \emph{Side effects of aspirin include cramping and stomach pain, and pain in the chest}. There are countless possible variants which may all communicate the correct answer; we provide additional real examples in the Appendix \ref{appendix:evaluations}. The flexibility of language means that parsing out the structured result to compare it to a reference (to calculate standard metrics like precision, recall, and F-1) is a non-trivial problem. This is in stark contrast to traditional approaches to tasks like NER and RE where models effectively classify input tokens instead of generating new ones from a vast vocabulary. 

Training models, either via traditional supervised learning or in-context few-shot learning, encourages models to comport with the structure of training instances.
We therefore focus our analysis on such supervised settings in this work, starting with an evaluation of few-shot learning with GPT-3 for RE. 
Nonetheless, even when supervised, LLMs used for RE are prone to generating outputs which may be accurate but nonetheless differ from the target. 
To address this, we enlist human annotators to judge whether the model outputs convey the same information as the reference targets.




\begin{table*}
\renewcommand*{\arraystretch}{1.5}
\small
\centering
\begin{tabular}{p{3cm}p{5.7cm}crrr}
\hline
\multicolumn{1}{c}{}                                                         & \multicolumn{1}{c}{\textbf{Method}}                                                                                    & \textbf{Params} & \multicolumn{1}{c}{\textbf{CONLL}} & \multicolumn{1}{c}{\textbf{ADE}} & \multicolumn{1}{c}{\textbf{NYT}} \\ \hline
\multirow{7}{*}{\texttt{\textbf{1. }}Fully supervised}                                            & \texttt{a.} SpERT* \cite{DBLP:journals/corr/abs-1909-07755}                                                                                                    & 110M          & $71.54$                         & $79.22$                       & -                       \\
                                                                             & \texttt{b.}  TANL  \cite{paolini2021structured}                                                                                                        & 220M          & $71.48$                         & $80.61$                       & $90.83$                       \\
                                                                             & \texttt{c.}  TANL (MT) \cite{paolini2021structured}                                                                                            & 220M          & $72.66$                         & $80.00$                       & $90.52$                       \\
                                                                             & \texttt{d.}  REBEL \cite{huguet-cabot-navigli-2021-rebel-relation}                                                                                           & 460M          & $75.44$                         & $82.21$                       & $92.00$                       \\
                                                                             & \texttt{e.}  Flan T5 (Large) \cite{https://doi.org/10.48550/arxiv.2210.11416}                                                                                                & 760M          & $75.28$                         & $83.15$                       & $91.03$                       \\ 
                                                                             & \texttt{f.}   $+$ \textbf{GPT-3-generated \emph{CoT}}                                                                                                & 760M          & \textbf{80.76}                        & \textbf{92.17}                       & \textbf{95.23}                       \\ \hline
\multirow{4}{*}{\texttt{\textbf{2. }}Few-shot}               & \texttt{a.} In-Context GPT-3      \cite{NEURIPS2020_1457c0d6}                                                                                        & 175B          & $76.53$                         & $82.66$                       & $61.79$                       \\ 
              & \texttt{b.} $+$ \emph{CoT}                                                                                        & 175B          & $78.18$                         & -                      & -                       \\        & c. Flan T5 (Large) w/ CoT Explanations \emph{and} reference labels generated from GPT-3 & 760M          & $76.13$                         & -                       & -                       \\ \hline
\end{tabular}
\caption{Comparison of (micro-F1) performance with recent generative (except SpERT) approaches in RE. Relation triplets/pairs are considered correct only if both of the corresponding entity types are correctly generated.}
\label{tab:results}
\end{table*}

\section{In-Context Few-Shot Learning with GPT-3 for RE}
\label{section:gpt3}

In this section we first describe our few-shot prompting strategy for GPT-3, and report the results realized by this approach across a set of RE corpora. 
 We adopt forms of instructional in-context few-shot prompting to GPT-3.\footnote{We provide details on the costs incurred for each of these experiments in the Appendix \ref{appendix:costs}.}
 Motivated by the preceding discussion regarding evaluation challenges, we collect human annotations judging the model's generations against the gold references. 
 Finally, using these annotations
 we report results achieved using GPT-3 with few-shot prompting for RE (Table \ref{tab:results}).
 All references to GPT-3 in this work refer to the ``text-davinci-002'' variant.

\subsection{Prompts}

We describe the prompts we use for each of the datasets considered in turn.

 

\paragraph{ADE} To construct prompts for ADE, we use the instructional prompt: \emph{List all (drug: adverse effects) pairs in the following text}, followed by an input text. 
We then select 12 examples (``shots'') at random from the training set, and for each we append the corresponding input followed by linearized target relations to the instructional prompt; this yields a prompt featuring 12 examples, comprising 755 tokens.
To make a prediction for a new example we append one last \emph{List all (drug: adverse effects) pairs in the following text} instruction followed by the corresponding text and then ask GPT-3 to generate text conditioned on this final prefix.
Specifically, we perform this generation using default parameters save for sampling temperature, which we set to 0.5.\footnote{In preliminary manual assessments, this seemed to yield qualitatively better outputs here than the default temperature.} We impose a maximum output length of 256 tokens. 




\paragraph{CoNLL} As an instructional prefix for CoNLL, we use: \emph{List the entities of the types [LOCATION, ORGANIZATION, PERSON] and relations of types [Organization Based In, Work For, Located In, Live In, Kill] among the entities in the given text}.
Since CoNLL is composed of four entity and five relation types, we constructed our prompt manually to contain at least one example of each entity and each relation type, for a total of 12 exemplars in the prompt. The total length of the CoNLL prompt was 960 tokens. To ensure fair comparison to prior work on generative RE over CoNLL, we use the same validation set as \citet{Eberts2019SpanbasedJE}. 






\paragraph{NYT} The large number of relations ($24$ in total) in the NYT dataset precludes the possibility of providing detailed instructions enumerating all entity and relation types. 
We instead shorten the instructional prefix by removing specific relation-type descriptors and create a prompt with only 20 exemplars capturing all entity and relation types. The size of this prompt was 2095 tokens. 




\paragraph{} We next aim to evaluate the performance of GPT-3 for RE when provided the above prompts. 
But doing so requires addressing the challenges inherent to evaluating LLMs for RE outlined above (and in prior work; \citealt{taille-etal-2020-lets}). 

\subsection{Manually re-evaluating ``errors''}

We quantify the errors in evaluation that occur when one uses ``strict'' measures of performance while using few-shot prompted LLMs for RE across each dataset.
We do this by acquiring human annotations (collected via Mechanical Turk; details in Appendix \ref{appendix:evaluations}) on model outputs, with respect to reference labels provided in the accompanying datasets. 
In particular, we show annotators ostensible ``false positive'' and ``false negative'' outputs produced by GPT-3 for these corpora---as would be computed using exact matching against references---and ask them to judge whether these are accurately categorized. 

On \textbf{ADE} we find that $51.67\%$ of ``false positives''---a slight majority---are more accurately viewed as \emph{true} positives, and $32.61\%$ of ``false negatives'' are deemed as, in fact, true negatives.
On \textbf{CoNLL} outputs, annotators marked $50.27\%$ of ``false positives'' as valid, and $36.6\%$ of ``false negatives'' as being accurate. 

As mentioned above, we were unable to design a prompt for \textbf{NYT} that yielded reasonable few-shot results with GPT-3. 
So we instead ask annotators to evaluate outputs from Flan-T5 fine-tuned on the NYT train set. 
In this case, they deemed $36.9\%$ and $22.97\%$ of ``false positives'' and ``false negatives'', respectively, to in fact be accurate. We present some illustrative cases in Figure \ref{fig:examples} and additional examples in Appendix Tables \ref{tab:fp} and \ref{tab:fn}. 




These findings imply that strict (exact-matching) evaluation against references for RE will be inaccurate (and pessimistic).
In the results we later report for LLMs, we therefore take into account these manual assessments.\footnote{One could also \emph{train} a model on manual assessments of ``false positives'' and ``false negatives'' to semi-automate this evaluation (avoiding the need to collect such judgments on entire testing sets); we provide results showing the feasibility of doing so in the Appendix \ref{appendix:evaluations}.}

\begin{figure*}[t]
    \centering
\includegraphics[scale=0.50]{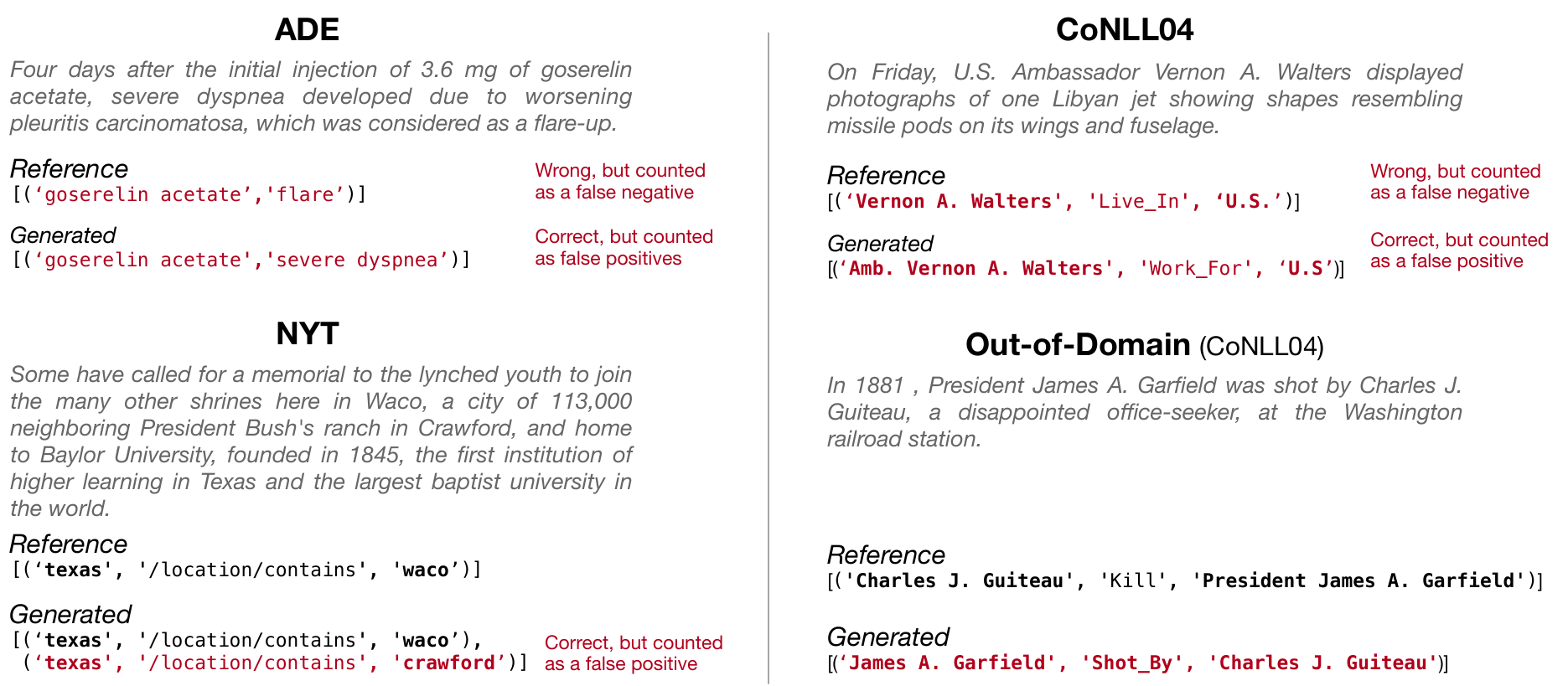}
\caption{Examples of misclassified FPs and FNs from GPT-3 (generated under few-shot in-context prompting scheme) under traditional evaluation of generative output. In each instance, the entity-type of \texttt{subject} and \texttt{object} was correctly identified. }
\label{fig:examples}
\end{figure*}


\subsection{Results} 

Using the above prompts and manual annotation process just described, we find that in most cases {\bf GPT-3 performs comparably to current \emph{fully supervised} SOTA RE models without fine-tuning and given only 12-20 training examples}. This can be seen in Table \ref{tab:results} (2.a).
We also find a substantial number of instances where the model correctly identifies relation pairs, which in fact are incorrectly marked in the references (detailed below in Section \ref{appendix:evaluations}). We observe additional issues with the NYT and CoNLL datasets which we discuss below.

\paragraph{CoNLL}
We find a number of relation triplets where the output does not conform to the set of valid relation types ($\sim$\% of relation triplets in the validation set).
Examining these triplets, we often find the out-of-domain relation-types to be either closely related to a correct CoNLL relation-type (e.g., \textit{shoot}$\longrightarrow$\textit{kill}) or otherwise correct even if not related to a CoNLL relation-type. 
There were a total of 18 input validation instances in which at least one of the generated relation triplet did not conform to a valid CoNLL relation; we provide a full list of these instances and the generated relation triplets in the Appendix \ref{appendix:evaluationsconlllist}. 

\paragraph{NYT} We find the strategy of omitting the relation descriptions in the prompt to be detrimental to the model's performance.
Contrary to our findings in ADE and CONLL, we observe a \textit{sharp decline} in Micro-F1 scores in case of NYT ($\sim$30 point reduction) as compared to the fully supervised SOTA. 
Further, we observe a non-trivial number of invalid or empty output instances ($\sim$10.6$\%$ of all generated sequences). 
These results highlight a remaining limitation of in-context learning with large language models: for datasets with long texts or a large number of targets, it is not possible to fit detailed instructions in the prompt. In light of the issues we were unable to evaluate this approach on the DocRED dataset, which we leave for future work. In such cases, traditional fine-tuning is the practical option. 

\vspace{0.25em}
Despite these limitations, the fact that GPT-3 is able to (marginally) outperform the current SOTA with in-context learning from tens of examples is encouraging. 
But GPT-3 is a massive opaque model available only via OpenAI's API (at cost). 
Further, fine-tuning GPT-3 would incur additional cost, and one would have access to the resultant model only via the OpenAI interface.
For these reasons, smaller, open-source LLMs for RE would be preferable. 
Next we show that by enriching supervision with \emph{Chain-of-Thought} (CoT) outputs elicited from GPT-3, we can achieve SOTA performance using Flan-T5 (Large).


\begin{figure*}
\centering
  \includegraphics[scale=0.7]{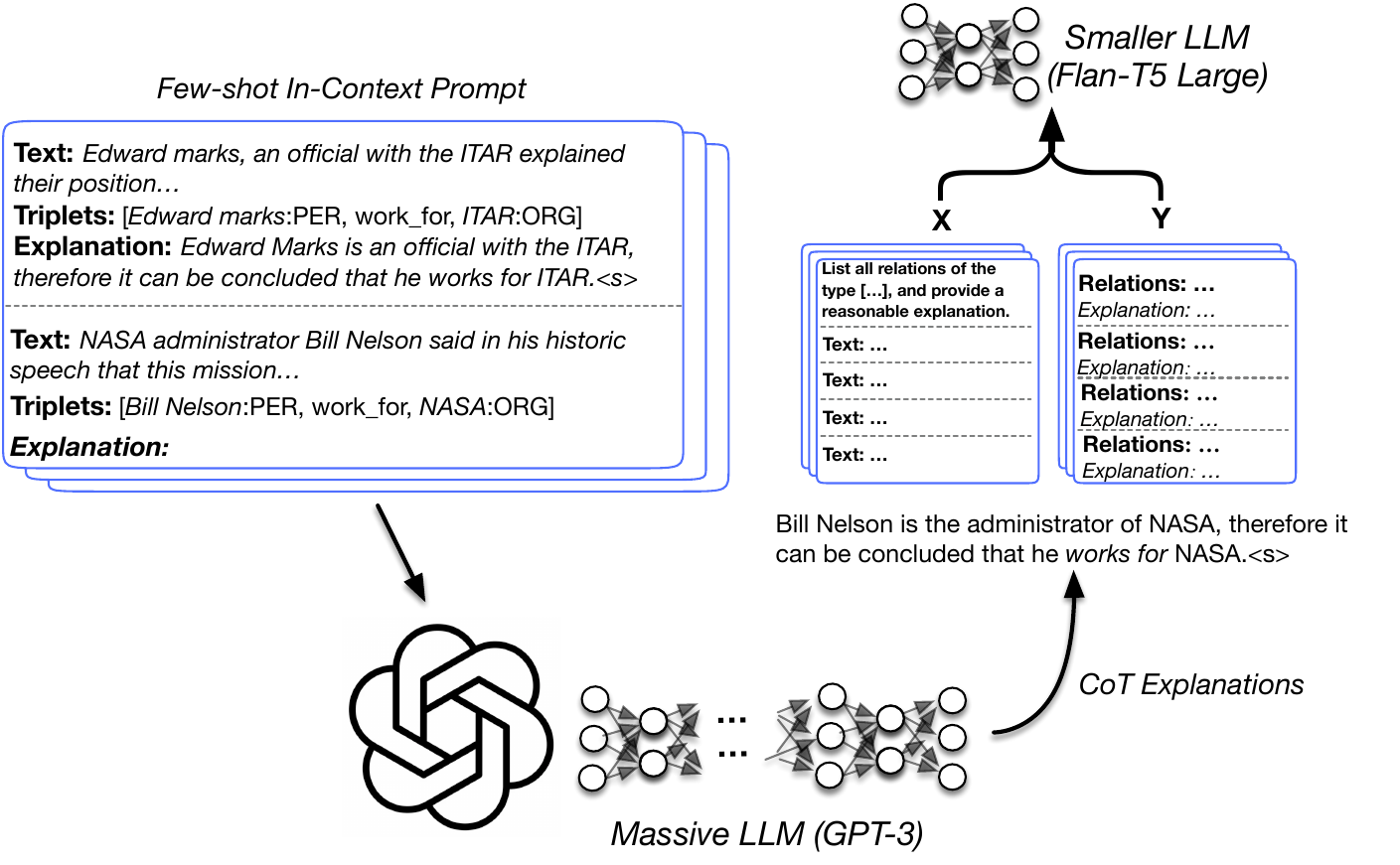}
  \caption{We propose fine-tuning Flan-T5 (large) for relation extraction (RE) using standard supervision \emph{and Chain-of-Thought (CoT) reasoning} elicited from GPT-3 for RE. This yields SOTA performance across all datasets considered, often by substantial margin ($\sim$5 points absolute gain in F1).}
  \label{fig:distil}
\end{figure*}

\section{SOTA RE Performance with Flan-T5} 
\label{section:flan}

We use Flan-T5 (Large), an LLM trained on a large number of tasks with instructional prompts. 
We first evaluate this in a few-shot setting (Section \ref{subsec:vanillapromptingflan}), shortening prompts in light of T5's smaller size, compared to GPT-3. 
We then consider fine-tuned variants, including a novel approach in which we train Flan-T5 using \emph{chain-of-thought} (CoT) style explanations for RE elicited from GPT-3. 
The latter strategy yields SOTA results across all datasets considered. 
 

\subsection{Few-Shot RE with Flan-T5}
\label{subsec:vanillapromptingflan}

For few-shot learning with Flan-T5, we use the same instructional prefixes (with examples) as we did for GPT-3 above, but we reduce the number of exemplars in the prompts to make them more concise.
We summarize our findings from these experiments on ADE and CoNLL below, and provide a full set of results in Appendix \ref{appendix:models}.

\vspace{0.25em}
\noindent \textbf{ADE} We include 7 (instead of the 12 used for GPT-3) randomly selected in-context examples for ADE. 
We observe a significant increase in non-conforming relation pairs in outputs ($13.9\%$ of generations). 
These often include outputs where the model generates the same token (or a set of tokens) repeatedly, or where relation tuples contain greater or fewer than 2 entities.
Unsurprisingly given these qualitative impressions, the model fares poorly under strict evaluation on the validation set, resulting in a $\sim20$ drop in F1 score compared to GPT-3. 

\vspace{0.25em}
\noindent \textbf{CoNLL} The prompt for CONLL consisted of 7 (in place of the 12 for GPT-3) exemplars inserted into the instructional prefix described above. 
Again we found that Flan-T5 generated many non-conforming outputs ($12.5\%$).
Additionally, we find that 
Flan-T5 generates a large number of out-of-domain relations between entities (over 120 unique relations), most of which are unrelated to CoNLL, making it impossible to meaningfully evaluate outputs (details in Appendix \ref{appendix:evaluations}). 

\vspace{0.25em}
\noindent \textbf{NYT} We exclude this dataset given the large set of relation and entity types, which---as discussed above---makes designing a prompt with sufficient instructions that also fits within the in-context window impossible. 
(We address this below via fine-tuning, which sidesteps the issue.)

\vspace{0.25em}

These results indicate that few-shot learning with Flan-T5 is not competitive with GPT-3, and so is not comparable to SOTA RE models. 
However, we next show that fine-tuning Flan-T5 can yield substantially better results, especially if one includes \emph{reasoning} about RE in the supervision.




\subsection{Fine-tuning Flan-T5 for RE}
\label{subsec:genexplanations}

We first perform standard fine-tuning for Flan-T5 (Large) using available training datasets.
We report results from the test set in Table \ref{tab:results} (1.e.). 
This yields performance equivalent to, but not better than, existing fully supervised models such as REBEL. 

As a potential mechanism to improve the performance of Flan-T5 for RE, we propose enriching the supervision used to fine-tune the model with \emph{chain-of-thought} (CoT; \citealt{wei2022chain}) explanations, which we elicit automatically from GPT-3 over the training instances.
Specifically, we craft a handful of such reasoning chains describing how target relations can be derived from the input texts. 
We provide the following three illustrative examples below.


\begin{flushleft}
\texttt{\textbf{Example Input (ADE) }To describe a case of severe skin necrosis resulting from peripheral intravenous administration of low-dose vasopressin in a patient with catecholamine-resistant septic shock.}

\texttt{\textbf{Target }[(vasopressin, skin necrosis)]   }

\texttt{\textbf{\textit{Explanation}} A case of skin necrosis was described after administration of low-dose vasopressin.}
\end{flushleft}

\begin{flushleft}
\texttt{\textbf{Example Input (CONLL) }In Colorado , 13 inches of snow in Denver Wednesday prompted officials to close Interstate 270 temporarily.}

\texttt{\textbf{Target }[(Denver, `Located In', Colorado)]}

\texttt{\textbf{\textit{Explanation}} - Denver officials closed Interstate 270 in Colorado, consequently we can see that Denver is located in Colorado.}
\end{flushleft}

\begin{flushleft}
\texttt{\textbf{Example Input (NYT) }It will be the final movie credited to Debra Hill, a film producer and native of Haddonfield, who produced ``Halloween'' and was considered a pioneering woman in film.}

\texttt{\textbf{Target } [[Debra Hill:Per, `place-of-birth', Haddonfield:Loc]]}

\texttt{\textbf{\textit{Explanation}} - Debra Hill was a film producer born (native of) in Haddonfield.}
\end{flushleft}


Next we evaluate the impact of CoT explanations in two settings: As additional context for prompting GPT-3, and then as additional supervision signal with which to train Flan-T5. 


\subsubsection{Eliciting CoT reasoning for RE}
We use the same prompts from the few-shot experiments above but augment them with CoT-style explanations (one per shot) written by one of the authors.
This yields moderate gains in the overall performance for GPT-3 ($\sim$3 and $\sim$2.2 micro-F1 points for ADE and CONLL, respectively; Table \ref{tab:results} 2.b),
and also reduces the number of non-conforming relations generated  
(from 13.9$\%$ to 0.8$\%$ on ADE, and from $12.5\%$ to 1.1$\%$ on CONLL).
Further, using CoT results in only one instance of an out-of-domain relation-type generated on CoNLL, compared to over 120 relations generated without CoT explanations.
In sum: using CoT in few-shot learning for RE with GPT-3 yields more standardized outputs, but does not much improve performance.
Next we propose to capitalize on CoTs automatically generated over training sets to enrich the supervision with which we train Flan-T5.

\subsubsection{Fine-tuning Flan-T5 with CoT explanations} 
\label{subsec:finetuneexplanations}

We augment target relations used to train Flan-T5 with CoT strings automatically generated by GPT-3 over the training dataset. 
Specifically, we modify the prompt used in Section \ref{section:gpt3} to generate \textit{CoT-style explanations} conditioned on the input \textit{and} relation reference labels.
The following is an example of the prompt we provide GPT-3 to elicit a \textit{CoT-explanation}:
\begin{flushleft}
\texttt{\textbf{Text: }This April 14 is the 125th anniversary of the night when Lincoln, the 16th president, was assassinated by John Wilkes Booth in the presidential box at Ford's Theatre.}

\texttt{\textbf{Target }[(John Wilkes Booth, `Kill', Lincoln)]}

\texttt{\textbf{\textit{Explanation}} - John Wilkes Booth assassinated Lincoln at the ford theatre.<s>}
\end{flushleft}

\begin{flushleft}
\texttt{\textbf{Text: }Ray is being held in Tennessee 's Brushy Mountain State Prison on a 99-year sentence for the April 4, 1968, slaying of King.}

\texttt{\textbf{Target } [[Ray, `Kill', King]]}

\texttt{\textbf{\textit{Explanation}} - }
\end{flushleft}

We then use these explanations along with reference relation labels as targets to fine-tune Flan-T5 (Large), as depicted in Figure \ref{fig:distil}. Overall, we found this strategy to be effective obtaining state-of-the-art results across datasets, while being much faster to train compared with existing fully supervised models. We summarize our findings below, and report results in Table \ref{tab:datastats} (1.f.).

\vspace{0.4em}
\noindent \textbf{ADE} We obtain explanations for the entire training set and fine-tune Flan-T5 Large with an instructional prefix with a batch size of 8, learning rate 3e-5 for 6 epochs. The dataset defines 10 folds of train/test splits, and we evaluate using the best checkpoint for each fold in the dataset. Our model yields a 9.97 point gain in micro F-1 score (averaged over the folds) over the existing fully supervised generative SOTA (REBEL; \citet{huguet-cabot-navigli-2021-rebel-relation}).


\vspace{0.4em}
\noindent \textbf{CONLL} For CONLL, we again obtain \textit{CoT-style} explanations for the entire dataset via GPT-3. We then fine-tune with a batch size of 4 and learning rate 3e-5  
for 10 epochs and evaluate using the best-performing checkpoint on the validation set. 
We see a 5.42 absolute point gain on the micro-F1 score over the existing fully-supervised generative SOTA.


\vspace{0.4em}
\noindent \textbf{NYT} comprises 56k training examples. In this case we generate CoT explanations via GPT-3 for only a subset of 25k examples (about half of the train set), due to its large size and the  associated cost.  
We fine-tune the model with a batch size of 4, learning rate 2e-5 for 4 epochs 
and then evaluate using the best performing checkpoint on the validation set. 
We obtain a 3.37 point gain on the micro-F1 score over the existing fully-supervised SOTA. 

\vspace{0.5em}
In sum, {\bf fine-tuning Flan-T5 (large) with both train labels and CoT explanations produced by GPT-3 yields SOTA performance across RE datasets by a considerable (5-10 points micro-F1) margin} (Figure \ref{fig:conllres}).

\subsubsection{``Fully Supervising'' Flan with GPT-3} 

Above we showed that Flan-T5 (large) outperforms existing RE methods by substantial margins when trained using CoTs from GPT-3.
Now we ask whether we can take this approach of distillation from GPT-3 even further by eliciting \emph{both labels and CoT explanations} from GPT-3 in a few-shot setting, and then using these to train Flan-T5. 
That is, above we used the reference labels for training, whereas here we use ``labels'' produced by GPT-3 given just a handful (10s) of training instances as shots.
We run this experiment only on CoNLL due to the cost of processing datasets in this way (which requires running few shot inference in GPT-3 over entire \emph{training} sets).

To generate the targets in this case, we start with an instructional prefix and 12 training instances from CoNLL and their corresponding human-written explanations; this is the same setup as the in-context GPT-3 model (Table \ref{tab:datastats} 2.b.), though here we apply this to the training instances. 
We then prompt GPT-3 on all training instances except for the 12 shots to produce pseudo labels (relations) and associated CoT explanations. 

Using this new \textit{GPT-generated training data}, we again fine-tune Flan-T5 (Large) as described above (Section \ref{subsec:finetuneexplanations}), and evaluate it on the validation set. 
This approach marginally outperforms the existing fully-supervised SOTA \cite{huguet-cabot-navigli-2021-rebel-relation}, but underperforms fine-tuning Flan with references references and GPT-generated explanations (Table \ref{tab:results}, 2.c.). 

\section{Related work}
\label{section:related-work}
Standard NLP methods for identifying relations in free text have included Conditional Random Fields \cite{10.5555/645530.655813}, structured SVMs \cite{10.1145/1015330.1015341}, and more recently, training large deep learning models with a joint objective \cite{eberts-ulges-2021-end, Eberts2019SpanbasedJE, wang-lu-2020-two} to identify entities and relations simultaneously. 
More recently, the rise of 
massive language models \cite{Radford2018ImprovingLU, Radford2019LanguageMA, NEURIPS2020_1457c0d6} has also 
motivated research into prompt-based learning methods for structured prediction \cite{wang-etal-2022-deepstruct}. 

\subsection{Relation extraction with pre-trained LMs}
\label{section:RE-with-LMs}
 
Several recently proposed RE approaches (which we have built upon here) have proposed addressing the task using conditional generative models to output string encodings---i.e., linearized forms---of target relations \cite{zeng-etal-2018-extracting, Zeng2020CopyMTLCM, Nayak2020EffectiveMO, huguet-cabot-navigli-2021-rebel-relation}.
\citet{paolini2021structured} proposed a framework that formulated many structured prediction tasks, including relation extraction, as a seq2seq problem where they decode outputs into structured information. \citet{huguet-cabot-navigli-2021-rebel-relation} extended this line of work by training a SOTA BART-style \cite{lewis-etal-2020-bart} model specifically for relation extraction using a unique triplet linearization strategy. 
Beyond these task-specific models, \citet{wang-etal-2022-deepstruct} proposed a task-agnostic structured pre-training scheme which enables zero-shot transfer to several structured prediction tasks. 

These past efforts focussed on \emph{solely} fine-tuning seq2seq models, adopting standard supervised approaches to learning to generate the relations expressed in a given input. 
({\tt REBEL} incorporated a pre-training scheme designed for RE \cite{huguet-cabot-navigli-2021-rebel-relation}, but this was in addition to a fine-tuning step.)
In this work we also evaluate the ability of large language models to perform \emph{few-shot} relation extraction via in-context learning; to our knowledge this is the first such evaluation for RE specifically, although few-shot learning more generally is an active sub-area of research. 


\subsection{Few Shot In-Context Learning}

Few shot in-context learning 
entails incorporating a few training examples into model prompts, effectively ``learning'' via the activations induced by passing these examples through the network at inference time. 
This has the advantage of completely forgoing model weight updates, which can be costly for LLMs \cite{wang-etal-2021-want-reduce}. An active area of research concerns such cross-task generalization capabilities \cite{ye-etal-2021-crossfit, Wei2022FinetunedLM, min-etal-2022-metaicl, DBLP:journals/corr/abs-2201-06910} of LLMs where a model learns a new, previously-unseen task efficiently with just a few examples. \citet{chen-etal-2022-improving} also proposed a self-supervised objective as an intermediate stage between pre-training and downstream few-shot learning. Recent work on few shot in-context learning has largely focused on the selection \cite{liu-etal-2022-makes} and ordering \cite{lu-etal-2022-fantastically} of exemplars included in the prompt 
provided to the model. 


\section{Conclusions and Future Directions}
\label{section:conclusions}

We have evaluated the capabilities of modern large language models (LLMs)---specifically GPT-3 and Flan T5 (Large)---on the task of Relation Extraction (RE). We found that, when evaluated carefully, GPT-3 performs comparably to fully supervised state-of-the-art (SOTA) models, given only 10s of examples. 
We then proposed a distillation technique in which we augmented target RE labels with \emph{Chain of Thought} (CoT) style explanations elicited from GPT-3 and used this to fine-tune Flan-T5; this yielded SOTA performance across all datasets considered, often by wide margins (5-10 points in F1).
Our results suggest that where feasible, LLMs should be a standard baseline for RE. 

\paragraph{Future directions} We have left several avenues open for further exploration. For example, evaluating LLMs like GPT-3 for RE required collecting manual annotations to identify ostensible ``false positive'' and ``false negative'' model outputs which were in fact accurate. 
Designing models to automate this evaluation might provide similar reliability without the accompanying costs; we provide preliminary work in this direction through the use of simple BERT-style classifiers in Appendix \ref{appendix:evaluations}.

\paragraph{\textcolor{red}{Brief update post-ACL acceptance:}} 

Owing to growing concerns about data contamination with closed LLMs including the recent variants of OpenAI's GPT\footnote{https://hitz-zentroa.github.io/lm-contamination/blog/}, we conducted additional experiments with a set of examples collected after the cutoff date for GPT-3.5's pretraining data (September 2021).
We collected 200 news headlines from Reuters\footnote{https://apify.com/zuzka/reuters-scraper} between June 1, 2023 - June 3, 2023 and evaluated the generalization of both GPT-3 and our fine-tuned Flan-T5 model on a CONLL-style RE task.
We used the same few-shot setup described in Section xx and measured precision on the new unseen examples via human annotations (on June 4, 2023). We observed that precision decreased $\sim$16 points for GPT-3 (from 78.31 to 62.01) and $\sim$7 points for Flan-T5 (from 81.22 to 74.70).
Even though we observe a sharper decline in performance with GPT-3, we were unable to
determine the main cause
for this drop in performance on the new data --- domain shift or data leakage. We leave a more detailed analysis of this issue for future work.


\section*{Limitations}
We have demonstrated that across three standard RE datasets, LLMs achieve SOTA results.
In particular, GPT-3 yields such performance even given only 10s of training sample for in-context learning.
We then showed that we can similarly achieve SOTA performance with the much smaller (and open-source) Flan T5 (Large) model, when trained using CoT generations produced by GPT-3.
We also highlighted key challenges for evaluation in this setting. 

But there are important limitations to these contributions. 
First, here we considered three standard RE datasets with binary relations but---as we discussed---we excluded more complex RE datasets.
For example, we did not consider corpora containing $n$-ary relations between entities~\cite{10.1093/nar/gkq906}.
We were also unable to run experiments on datasets with lengthy texts and a large number of relations, such as DocRED \cite{yao-etal-2021-codred}, due to the necessary prompt lengths for such inputs.

Second, while we found that CoT-style explanations generated by GPT-3 can be fruitfully used as additional supervision to fine-tune smaller language models, we made no attempt to evaluate the \textit{quality} of these generated explanations which may have an impact on the model performance. 

Third, we did not fine-tune GPT-3 on the RE datasets, mainly  due to the cost of doing so.
It is likely that a fine-tuned GPT-3 would yield performance superior to the results we achieved with Flan T5 (which constitute current SOTA).
But, in addition to the costs necessary for fine-tuning this model, the resultant weights would not be accessible to run locally in any case; one would have access to it only via the OpenAI interface, which motivated our decision to fine-tune the smaller and open-source Flan T5 instead.

Finally, we \textit{only} experiment with datasets curated in the English language and therefore, we do not know that the issues we have highlighted could replicate in the same way in other languages.

\section*{Ethics Statement}
Our work required an extensive manual annotation and evaluation process which involved using Amazon Mechanical Turk.
 Turk requires we pay workers \textit{per annotation}, so we have to estimate the time required for each task.
To do so, we (the authors) carried out a small number of these annotations ourselves to determine fair approximate hourly compensation. We then set the price per annotation such that it averages out to $\$15$/hour (we pay this rate irrespective of geographic location of the workers). We also provided our recruited AMT workers 20\% additional time per annotation.

\section*{Acknowledgements}

This work was supported in part by the National Institutes of Health (NIH) under the National Library of Medicine (NLM) grant R01LM012086 and by the National Science Foundation (NSF) grant III-1750978. 
\bibliography{anthology,custom}

\begin{thebibliography}{36}
\expandafter\ifx\csname natexlab\endcsname\relax\def\natexlab#1{#1}\fi

\bibitem[{Brown et~al.(2020{\natexlab{a}})Brown, Mann, Ryder, Subbiah, Kaplan,
  Dhariwal, Neelakantan, Shyam, Sastry, Askell, Agarwal, Herbert-Voss, Krueger,
  Henighan, Child, Ramesh, Ziegler, Wu, Winter, Hesse, Chen, Sigler, Litwin,
  Gray, Chess, Clark, Berner, McCandlish, Radford, Sutskever, and
  Amodei}]{NEURIPS2020_1457c0d6}
Tom Brown, Benjamin Mann, Nick Ryder, Melanie Subbiah, Jared~D Kaplan, Prafulla
  Dhariwal, Arvind Neelakantan, Pranav Shyam, Girish Sastry, Amanda Askell,
  Sandhini Agarwal, Ariel Herbert-Voss, Gretchen Krueger, Tom Henighan, Rewon
  Child, Aditya Ramesh, Daniel Ziegler, Jeffrey Wu, Clemens Winter, Chris
  Hesse, Mark Chen, Eric Sigler, Mateusz Litwin, Scott Gray, Benjamin Chess,
  Jack Clark, Christopher Berner, Sam McCandlish, Alec Radford, Ilya Sutskever,
  and Dario Amodei. 2020{\natexlab{a}}.
\newblock \href
  {https://proceedings.neurips.cc/paper/2020/file/1457c0d6bfcb4967418bfb8ac142f64a-Paper.pdf}
  {Language models are few-shot learners}.
\newblock In \emph{Advances in Neural Information Processing Systems},
  volume~33, pages 1877--1901. Curran Associates, Inc.

\bibitem[{Brown et~al.(2020{\natexlab{b}})Brown, Mann, Ryder, Subbiah, Kaplan,
  Dhariwal, Neelakantan, Shyam, Sastry, Askell, Agarwal, Herbert-Voss, Krueger,
  Henighan, Child, Ramesh, Ziegler, Wu, Winter, Hesse, Chen, Sigler, Litwin,
  Gray, Chess, Clark, Berner, McCandlish, Radford, Sutskever, and
  Amodei}]{Brown2020LanguageMA}
Tom~B. Brown, Benjamin Mann, Nick Ryder, Melanie Subbiah, Jared Kaplan,
  Prafulla Dhariwal, Arvind Neelakantan, Pranav Shyam, Girish Sastry, Amanda
  Askell, Sandhini Agarwal, Ariel Herbert-Voss, Gretchen Krueger, T.~J.
  Henighan, Rewon Child, Aditya Ramesh, Daniel~M. Ziegler, Jeff Wu, Clemens
  Winter, Christopher Hesse, Mark Chen, Eric Sigler, Mateusz Litwin, Scott
  Gray, Benjamin Chess, Jack Clark, Christopher Berner, Sam McCandlish, Alec
  Radford, Ilya Sutskever, and Dario Amodei. 2020{\natexlab{b}}.
\newblock Language models are few-shot learners.
\newblock \emph{ArXiv}, abs/2005.14165.

\bibitem[{Chen et~al.(2022)Chen, Du, Pasunuru, Mihaylov, Iyer, Stoyanov, and
  Kozareva}]{chen-etal-2022-improving}
Mingda Chen, Jingfei Du, Ramakanth Pasunuru, Todor Mihaylov, Srini Iyer,
  Veselin Stoyanov, and Zornitsa Kozareva. 2022.
\newblock \href {https://doi.org/10.18653/v1/2022.naacl-main.260} {Improving
  in-context few-shot learning via self-supervised training}.
\newblock In \emph{Proceedings of the 2022 Conference of the North American
  Chapter of the Association for Computational Linguistics: Human Language
  Technologies}, pages 3558--3573, Seattle, United States. Association for
  Computational Linguistics.

\bibitem[{Chung et~al.(2022)Chung, Hou, Longpre, Zoph, Tay, Fedus, Li, Wang,
  Dehghani, Brahma, Webson, Gu, Dai, Suzgun, Chen, Chowdhery, Castro-Ros,
  Pellat, Robinson, Valter, Narang, Mishra, Yu, Zhao, Huang, Dai, Yu, Petrov,
  Chi, Dean, Devlin, Roberts, Zhou, Le, and
  Wei}]{https://doi.org/10.48550/arxiv.2210.11416}
Hyung~Won Chung, Le~Hou, Shayne Longpre, Barret Zoph, Yi~Tay, William Fedus,
  Yunxuan Li, Xuezhi Wang, Mostafa Dehghani, Siddhartha Brahma, Albert Webson,
  Shixiang~Shane Gu, Zhuyun Dai, Mirac Suzgun, Xinyun Chen, Aakanksha
  Chowdhery, Alex Castro-Ros, Marie Pellat, Kevin Robinson, Dasha Valter,
  Sharan Narang, Gaurav Mishra, Adams Yu, Vincent Zhao, Yanping Huang, Andrew
  Dai, Hongkun Yu, Slav Petrov, Ed~H. Chi, Jeff Dean, Jacob Devlin, Adam
  Roberts, Denny Zhou, Quoc~V. Le, and Jason Wei. 2022.
\newblock \href {https://doi.org/10.48550/ARXIV.2210.11416} {Scaling
  instruction-finetuned language models}.

\bibitem[{Devlin et~al.(2019)Devlin, Chang, Lee, and
  Toutanova}]{devlin-etal-2019-bert}
Jacob Devlin, Ming-Wei Chang, Kenton Lee, and Kristina Toutanova. 2019.
\newblock \href {https://doi.org/10.18653/v1/N19-1423} {{BERT}: Pre-training of
  deep bidirectional transformers for language understanding}.
\newblock In \emph{Proceedings of the 2019 Conference of the North {A}merican
  Chapter of the Association for Computational Linguistics: Human Language
  Technologies, Volume 1 (Long and Short Papers)}, pages 4171--4186,
  Minneapolis, Minnesota. Association for Computational Linguistics.

\bibitem[{Eberts and Ulges(2019{\natexlab{a}})}]{Eberts2019SpanbasedJE}
Markus Eberts and Adrian Ulges. 2019{\natexlab{a}}.
\newblock Span-based joint entity and relation extraction with transformer
  pre-training.
\newblock \emph{ArXiv}, abs/1909.07755.

\bibitem[{Eberts and
  Ulges(2019{\natexlab{b}})}]{DBLP:journals/corr/abs-1909-07755}
Markus Eberts and Adrian Ulges. 2019{\natexlab{b}}.
\newblock \href {http://arxiv.org/abs/1909.07755} {Span-based joint entity and
  relation extraction with transformer pre-training}.
\newblock \emph{CoRR}, abs/1909.07755.

\bibitem[{Eberts and Ulges(2021)}]{eberts-ulges-2021-end}
Markus Eberts and Adrian Ulges. 2021.
\newblock \href {https://doi.org/10.18653/v1/2021.eacl-main.319} {An end-to-end
  model for entity-level relation extraction using multi-instance learning}.
\newblock In \emph{Proceedings of the 16th Conference of the European Chapter
  of the Association for Computational Linguistics: Main Volume}, pages
  3650--3660, Online. Association for Computational Linguistics.

\bibitem[{Gurulingappa et~al.(2012)Gurulingappa, Rajput, Roberts, Fluck,
  Hofmann-Apitius, and Toldo}]{Gurulingappa2012DevelopmentOA}
Harsha Gurulingappa, Abdul~Mateen Rajput, Angus Roberts, Juliane Fluck, Martin
  Hofmann-Apitius, and Luca Toldo. 2012.
\newblock Development of a benchmark corpus to support the automatic extraction
  of drug-related adverse effects from medical case reports.
\newblock \emph{Journal of biomedical informatics}, 45 5:885--92.

\bibitem[{Huguet~Cabot and
  Navigli(2021)}]{huguet-cabot-navigli-2021-rebel-relation}
Pere-Llu{\'\i}s Huguet~Cabot and Roberto Navigli. 2021.
\newblock \href {https://doi.org/10.18653/v1/2021.findings-emnlp.204} {{REBEL}:
  Relation extraction by end-to-end language generation}.
\newblock In \emph{Findings of the Association for Computational Linguistics:
  EMNLP 2021}, pages 2370--2381, Punta Cana, Dominican Republic. Association
  for Computational Linguistics.

\bibitem[{Lafferty et~al.(2001)Lafferty, McCallum, and
  Pereira}]{10.5555/645530.655813}
John~D. Lafferty, Andrew McCallum, and Fernando C.~N. Pereira. 2001.
\newblock Conditional random fields: Probabilistic models for segmenting and
  labeling sequence data.
\newblock In \emph{Proceedings of the Eighteenth International Conference on
  Machine Learning}, ICML '01, page 282–289, San Francisco, CA, USA. Morgan
  Kaufmann Publishers Inc.

\bibitem[{Lewis et~al.(2020)Lewis, Liu, Goyal, Ghazvininejad, Mohamed, Levy,
  Stoyanov, and Zettlemoyer}]{lewis-etal-2020-bart}
Mike Lewis, Yinhan Liu, Naman Goyal, Marjan Ghazvininejad, Abdelrahman Mohamed,
  Omer Levy, Veselin Stoyanov, and Luke Zettlemoyer. 2020.
\newblock \href {https://doi.org/10.18653/v1/2020.acl-main.703} {{BART}:
  Denoising sequence-to-sequence pre-training for natural language generation,
  translation, and comprehension}.
\newblock In \emph{Proceedings of the 58th Annual Meeting of the Association
  for Computational Linguistics}, pages 7871--7880, Online. Association for
  Computational Linguistics.

\bibitem[{Liu et~al.(2022)Liu, Shen, Zhang, Dolan, Carin, and
  Chen}]{liu-etal-2022-makes}
Jiachang Liu, Dinghan Shen, Yizhe Zhang, Bill Dolan, Lawrence Carin, and Weizhu
  Chen. 2022.
\newblock \href {https://doi.org/10.18653/v1/2022.deelio-1.10} {What makes good
  in-context examples for {GPT}-3?}
\newblock In \emph{Proceedings of Deep Learning Inside Out (DeeLIO 2022): The
  3rd Workshop on Knowledge Extraction and Integration for Deep Learning
  Architectures}, pages 100--114, Dublin, Ireland and Online. Association for
  Computational Linguistics.

\bibitem[{Lu et~al.(2022{\natexlab{a}})Lu, Bartolo, Moore, Riedel, and
  Stenetorp}]{lu-etal-2022-fantastically}
Yao Lu, Max Bartolo, Alastair Moore, Sebastian Riedel, and Pontus Stenetorp.
  2022{\natexlab{a}}.
\newblock \href {https://doi.org/10.18653/v1/2022.acl-long.556} {Fantastically
  ordered prompts and where to find them: Overcoming few-shot prompt order
  sensitivity}.
\newblock In \emph{Proceedings of the 60th Annual Meeting of the Association
  for Computational Linguistics (Volume 1: Long Papers)}, pages 8086--8098,
  Dublin, Ireland. Association for Computational Linguistics.

\bibitem[{Lu et~al.(2022{\natexlab{b}})Lu, Liu, Dai, Xiao, Lin, Han, Sun, and
  Wu}]{lu-etal-2022-unified}
Yaojie Lu, Qing Liu, Dai Dai, Xinyan Xiao, Hongyu Lin, Xianpei Han, Le~Sun, and
  Hua Wu. 2022{\natexlab{b}}.
\newblock \href {https://doi.org/10.18653/v1/2022.acl-long.395} {Unified
  structure generation for universal information extraction}.
\newblock In \emph{Proceedings of the 60th Annual Meeting of the Association
  for Computational Linguistics (Volume 1: Long Papers)}, pages 5755--5772,
  Dublin, Ireland. Association for Computational Linguistics.

\bibitem[{Min et~al.(2022)Min, Lewis, Zettlemoyer, and
  Hajishirzi}]{min-etal-2022-metaicl}
Sewon Min, Mike Lewis, Luke Zettlemoyer, and Hannaneh Hajishirzi. 2022.
\newblock \href {https://doi.org/10.18653/v1/2022.naacl-main.201} {{M}eta{ICL}:
  Learning to learn in context}.
\newblock In \emph{Proceedings of the 2022 Conference of the North American
  Chapter of the Association for Computational Linguistics: Human Language
  Technologies}, pages 2791--2809, Seattle, United States. Association for
  Computational Linguistics.

\bibitem[{Nayak and Ng(2020)}]{Nayak2020EffectiveMO}
Tapas Nayak and Hwee~Tou Ng. 2020.
\newblock Effective modeling of encoder-decoder architecture for joint entity
  and relation extraction.
\newblock In \emph{AAAI Conference on Artificial Intelligence}.

\bibitem[{Paolini et~al.(2021)Paolini, Athiwaratkun, Krone, Ma, Achille,
  ANUBHAI, dos Santos, Xiang, and Soatto}]{paolini2021structured}
Giovanni Paolini, Ben Athiwaratkun, Jason Krone, Jie Ma, Alessandro Achille,
  RISHITA ANUBHAI, Cicero~Nogueira dos Santos, Bing Xiang, and Stefano Soatto.
  2021.
\newblock \href {https://openreview.net/forum?id=US-TP-xnXI} {Structured
  prediction as translation between augmented natural languages}.
\newblock In \emph{International Conference on Learning Representations}.

\bibitem[{Radford and Narasimhan(2018)}]{Radford2018ImprovingLU}
Alec Radford and Karthik Narasimhan. 2018.
\newblock Improving language understanding by generative pre-training.

\bibitem[{Radford et~al.(2019)Radford, Wu, Child, Luan, Amodei, and
  Sutskever}]{Radford2019LanguageMA}
Alec Radford, Jeff Wu, Rewon Child, David Luan, Dario Amodei, and Ilya
  Sutskever. 2019.
\newblock Language models are unsupervised multitask learners.

\bibitem[{Riedel et~al.(2010)Riedel, Yao, and McCallum}]{Riedel2010ModelingRA}
Sebastian Riedel, Limin Yao, and Andrew McCallum. 2010.
\newblock Modeling relations and their mentions without labeled text.
\newblock In \emph{ECML/PKDD}.

\bibitem[{Roth and Yih(2004)}]{roth-yih-2004-linear}
Dan Roth and Wen-tau Yih. 2004.
\newblock \href {https://aclanthology.org/W04-2401} {A linear programming
  formulation for global inference in natural language tasks}.
\newblock In \emph{Proceedings of the Eighth Conference on Computational
  Natural Language Learning ({C}o{NLL}-2004) at {HLT}-{NAACL} 2004}, pages
  1--8, Boston, Massachusetts, USA. Association for Computational Linguistics.

\bibitem[{Taboureau et~al.(2010)Taboureau, Nielsen, Audouze, Weinhold,
  Edsgärd, Roque, Kouskoumvekaki, Bora, Curpan, Jensen, Brunak, and
  Oprea}]{10.1093/nar/gkq906}
Olivier Taboureau, Sonny~Kim Nielsen, Karine Audouze, Nils Weinhold, Daniel
  Edsgärd, Francisco~S. Roque, Irene Kouskoumvekaki, Alina Bora, Ramona
  Curpan, Thomas~Skøt Jensen, Søren Brunak, and Tudor~I. Oprea. 2010.
\newblock \href {https://doi.org/10.1093/nar/gkq906} {{ChemProt: a disease
  chemical biology database}}.
\newblock \emph{Nucleic Acids Research}, 39:D367--D372.

\bibitem[{Taill{\'e} et~al.(2020)Taill{\'e}, Guigue, Scoutheeten, and
  Gallinari}]{taille-etal-2020-lets}
Bruno Taill{\'e}, Vincent Guigue, Geoffrey Scoutheeten, and Patrick Gallinari.
  2020.
\newblock \href {https://doi.org/10.18653/v1/2020.emnlp-main.301} {Let{'}s
  {S}top {I}ncorrect {C}omparisons in {E}nd-to-end {R}elation {E}xtraction!}
\newblock In \emph{Proceedings of the 2020 Conference on Empirical Methods in
  Natural Language Processing (EMNLP)}, pages 3689--3701, Online. Association
  for Computational Linguistics.

\bibitem[{Tsochantaridis et~al.(2004)Tsochantaridis, Hofmann, Joachims, and
  Altun}]{10.1145/1015330.1015341}
Ioannis Tsochantaridis, Thomas Hofmann, Thorsten Joachims, and Yasemin Altun.
  2004.
\newblock \href {https://doi.org/10.1145/1015330.1015341} {Support vector
  machine learning for interdependent and structured output spaces}.
\newblock In \emph{Proceedings of the Twenty-First International Conference on
  Machine Learning}, ICML '04, page 104, New York, NY, USA. Association for
  Computing Machinery.

\bibitem[{Wang et~al.(2022)Wang, Liu, Chen, Hong, Tang, and
  Song}]{wang-etal-2022-deepstruct}
Chenguang Wang, Xiao Liu, Zui Chen, Haoyun Hong, Jie Tang, and Dawn Song. 2022.
\newblock \href {https://doi.org/10.18653/v1/2022.findings-acl.67}
  {{D}eep{S}truct: Pretraining of language models for structure prediction}.
\newblock In \emph{Findings of the Association for Computational Linguistics:
  ACL 2022}, pages 803--823, Dublin, Ireland. Association for Computational
  Linguistics.

\bibitem[{Wang and Lu(2020)}]{wang-lu-2020-two}
Jue Wang and Wei Lu. 2020.
\newblock \href {https://doi.org/10.18653/v1/2020.emnlp-main.133} {Two are
  better than one: Joint entity and relation extraction with table-sequence
  encoders}.
\newblock In \emph{Proceedings of the 2020 Conference on Empirical Methods in
  Natural Language Processing (EMNLP)}, pages 1706--1721, Online. Association
  for Computational Linguistics.

\bibitem[{Wang et~al.(2021)Wang, Liu, Xu, Zhu, and
  Zeng}]{wang-etal-2021-want-reduce}
Shuohang Wang, Yang Liu, Yichong Xu, Chenguang Zhu, and Michael Zeng. 2021.
\newblock \href {https://doi.org/10.18653/v1/2021.findings-emnlp.354} {Want to
  reduce labeling cost? {GPT}-3 can help}.
\newblock In \emph{Findings of the Association for Computational Linguistics:
  EMNLP 2021}, pages 4195--4205, Punta Cana, Dominican Republic. Association
  for Computational Linguistics.

\bibitem[{Wei et~al.(2022{\natexlab{a}})Wei, Bosma, Zhao, Guu, Yu, Lester, Du,
  Dai, and Le}]{Wei2022FinetunedLM}
Jason Wei, Maarten Bosma, Vincent Zhao, Kelvin Guu, Adams~Wei Yu, Brian Lester,
  Nan Du, Andrew~M. Dai, and Quoc~V. Le. 2022{\natexlab{a}}.
\newblock Finetuned language models are zero-shot learners.
\newblock \emph{ArXiv}, abs/2109.01652.

\bibitem[{Wei et~al.(2022{\natexlab{b}})Wei, Wang, Schuurmans, Bosma, Chi, Le,
  and Zhou}]{wei2022chain}
Jason Wei, Xuezhi Wang, Dale Schuurmans, Maarten Bosma, Ed~Chi, Quoc Le, and
  Denny Zhou. 2022{\natexlab{b}}.
\newblock Chain of thought prompting elicits reasoning in large language
  models.
\newblock \emph{arXiv preprint arXiv:2201.11903}.

\bibitem[{Xu et~al.(2022)Xu, Chen, Du, Shao, Wang, Li, and
  Yang}]{DBLP:journals/corr/abs-2201-06910}
Hanwei Xu, Yujun Chen, Yulun Du, Nan Shao, Yanggang Wang, Haiyu Li, and Zhilin
  Yang. 2022.
\newblock \href {http://arxiv.org/abs/2201.06910} {Zeroprompt: Scaling
  prompt-based pretraining to 1, 000 tasks improves zero-shot generalization}.
\newblock \emph{CoRR}, abs/2201.06910.

\bibitem[{Yao et~al.(2021)Yao, Du, Lin, Li, Liu, Zhou, and
  Sun}]{yao-etal-2021-codred}
Yuan Yao, Jiaju Du, Yankai Lin, Peng Li, Zhiyuan Liu, Jie Zhou, and Maosong
  Sun. 2021.
\newblock \href {https://doi.org/10.18653/v1/2021.emnlp-main.366} {{C}od{RED}:
  A cross-document relation extraction dataset for acquiring knowledge in the
  wild}.
\newblock In \emph{Proceedings of the 2021 Conference on Empirical Methods in
  Natural Language Processing}, pages 4452--4472, Online and Punta Cana,
  Dominican Republic. Association for Computational Linguistics.

\bibitem[{Yao et~al.(2019)Yao, Ye, Li, Han, Lin, Liu, Liu, Huang, Zhou, and
  Sun}]{yao-etal-2019-docred}
Yuan Yao, Deming Ye, Peng Li, Xu~Han, Yankai Lin, Zhenghao Liu, Zhiyuan Liu,
  Lixin Huang, Jie Zhou, and Maosong Sun. 2019.
\newblock \href {https://doi.org/10.18653/v1/P19-1074} {{D}oc{RED}: A
  large-scale document-level relation extraction dataset}.
\newblock In \emph{Proceedings of the 57th Annual Meeting of the Association
  for Computational Linguistics}, pages 764--777, Florence, Italy. Association
  for Computational Linguistics.

\bibitem[{Ye et~al.(2021)Ye, Lin, and Ren}]{ye-etal-2021-crossfit}
Qinyuan Ye, Bill~Yuchen Lin, and Xiang Ren. 2021.
\newblock \href {https://doi.org/10.18653/v1/2021.emnlp-main.572}
  {{C}ross{F}it: A few-shot learning challenge for cross-task generalization in
  {NLP}}.
\newblock In \emph{Proceedings of the 2021 Conference on Empirical Methods in
  Natural Language Processing}, pages 7163--7189, Online and Punta Cana,
  Dominican Republic. Association for Computational Linguistics.

\bibitem[{Zeng et~al.(2020)Zeng, Zhang, and Liu}]{Zeng2020CopyMTLCM}
Daojian Zeng, Haoran Zhang, and Qianying Liu. 2020.
\newblock Copymtl: Copy mechanism for joint extraction of entities and
  relations with multi-task learning.
\newblock \emph{ArXiv}, abs/1911.10438.

\bibitem[{Zeng et~al.(2018)Zeng, Zeng, He, Liu, and
  Zhao}]{zeng-etal-2018-extracting}
Xiangrong Zeng, Daojian Zeng, Shizhu He, Kang Liu, and Jun Zhao. 2018.
\newblock \href {https://doi.org/10.18653/v1/P18-1047} {Extracting relational
  facts by an end-to-end neural model with copy mechanism}.
\newblock In \emph{Proceedings of the 56th Annual Meeting of the Association
  for Computational Linguistics (Volume 1: Long Papers)}, pages 506--514,
  Melbourne, Australia. Association for Computational Linguistics.

\end{thebibliography}
\bibliographystyle{acl_natbib}

\clearpage

\appendix
\section{Datasets}
\label{appendix:datasets}
We considered and conducted the evaluation of our methods on the following datasets. 
Basic data statistics are also reported in Table \ref{tab:datastats}.

\paragraph{ADE}
Adverse Drug Events \cite{Gurulingappa2012DevelopmentOA} contains binary relations of (\texttt{drug}, \texttt{adverse event}) pairs. 
Drugs and adverse events are the only two entity types. This dataset provides a 10-fold split. 


\paragraph{CONLL04}
The CoNLL04 consists of sentences from news articles that were annotated for the mentioned entities and relations between entities~\cite{roth-yih-2004-linear}. It includes four entity types (\textit{PER, ORG, LOC, OTH}) and five possible relations (\textit{KILL, WORK$\_$FOR, LIVE$\_$IN, LOCATED$\_$IN, ORG$\_$BASED$\_$IN}). 


\paragraph{NYT}
The NYT comprises sentences sampled from New York Times news articles published between 1987 and 2007~\cite{Riedel2010ModelingRA}. The data was distantly annotated with relations triplets from FreeBase. We use a processed version of NYT \cite{zeng-etal-2018-extracting} containing three overlapping entity types (\textit{LOC, PER, ORG}) and 24 relation types. 

\paragraph{DocRED}
 Originally designed as a relation classification task, DocRED \cite{yao-etal-2019-docred} differs considerably from the other datasets considered in this work in two important ways: 
(1) It comprises long texts which feature relations between entities at a \textit{document-level}; (2) It contains annotations for 6 entity types and 96 relation types, with an average of 19.9 entities and 19.5 relation instances per document. 

\section{Models and Reproducibility}
\label{appendix:models}

\begin{table}[]
\centering
\scriptsize
\begin{tabular}{lcrrr}
\hline
\multicolumn{1}{c}{\textbf{Model}}                                                                                                & \textbf{Data}  & \multicolumn{1}{c}{\textbf{P}} & \multicolumn{1}{c}{\textbf{R}} & \multicolumn{1}{c}{\textbf{F-1}} \\ \hline
\multirow{3}{*}{\begin{tabular}[c]{@{}l@{}}Few-Shot In-Context\\ Prompting GPT-3\end{tabular}}                           & ADE   & 80.85                 & 84.54                 & 82.66                   \\
                                                                                                                         & CoNLL & 78.31                 & 74.82                 & 76.53                   \\
                                                                                                                         & NYT   & 66.63                 & 70.58                 & 68.55                   \\ \hline
\multirow{3}{*}{\begin{tabular}[c]{@{}l@{}}Vanilla Fine-Tune \\ Flan-T5-Large\end{tabular}}                              & ADE   & 89.11                 & 77.93                 & 83.15                   \\
                                                                                                                         & CoNLL & 78.81                 & 72.05                 & 75.28                   \\
                                                                                                                         & NYT   & 91.82                 & 90.25                 & 91.03                   \\ \hline
\multirow{3}{*}{\begin{tabular}[c]{@{}l@{}}Fine-Tune Flan\\ on GPT-3-generated CoT\end{tabular}}                         & ADE   & 91.74                 & 92.60                 & 92.17                   \\
                                                                                                                         & CoNLL & 81.22                 & 80.31                 & 80.76                   \\
                                                                                                                         & NYT   & 95.49                 & 94.97                 & 95.23                   \\ \hline
\begin{tabular}[c]{@{}l@{}}Fine-Tune Flan w/ CoT \\ Explanations and Reference \\ labels generated from GPT\end{tabular} & CoNLL & 76.41                 & 75.85                 & 76.13                   \\ \hline
\end{tabular}
\caption{Average micro metrics over 5 seeds for the test sets (10-folds for ADE). }
\label{tab:extrares}
\end{table}

\begin{table*}[]
\centering
\small
\begin{tabular}{@{}p{4cm}lccccc@{}}
\toprule
\multicolumn{1}{c}{\textbf{Model}}                                                                              & \multicolumn{1}{c}{\textbf{Data}} & \textbf{Batch Size} & \textbf{Warm-up} & \multicolumn{1}{l}{\textbf{Learning Rate}} & \begin{tabular}[c]{@{}c@{}}\textbf{Time/Epoch}\\ (minutes)\end{tabular} & \textbf{Max Epochs} \\ \midrule
\multirow{3}{*}{\begin{tabular}[c]{@{}l@{}}Vanilla Fine-Tune \\ Flan-T5-Large\end{tabular}}                                                        & ADE                      & 8          & 10\%    & 3e-5                              & 36                                                             & 6          \\
                                                                                                       & CoNLL                    & 4          & 12\%    & 3e-5                              & 22                                                             & 10         \\
                                                                                                       & NYT                      & 4          & 12\%    & 2e-5                              & 99                                                             & 4          \\ \midrule
\multirow{3}{*}{\begin{tabular}[c]{@{}l@{}}\textbf{Fine-Tune Flan on}\\ \textbf{GPT-3-generated CoT}\end{tabular}}                                                                 & ADE                      & 8          & 10\%    & 3e-5                              & 38                                                             & 6          \\
                                                                                                       & CoNLL                    & 4          & 12\%    & 3e-5                              & 28                                                             & 10         \\
                                                                                                       & NYT                      & 4          & 12\%    & 2e-5                              & 107                                                            & 4          \\ \midrule
\multirow{3}{*}{\begin{tabular}[c]{@{}l@{}}Fine-Tune Flan w/ CoT \\ Explanations and Reference \\ labels generated from GPT\end{tabular}} & ADE                      & 8          & 10\%    & 3e-5                              & 37                                                             & 6          \\
                                                                                                       & CoNLL                    & 4          & 12\%    & 3e-5                              & 28                                                             & 10         \\
                                                                                                       & NYT                      & 4          & 12\%    & 2e-5                              & 109                                                            & 4          \\ \bottomrule
\end{tabular}
\caption{Hyperparameters and compute time for the fully fine-tuned Flan models (corresponding to main results table \ref{tab:results}).}
\label{tab:training}
\end{table*}

We provide average micro metrics over 5 seeds across each dataset in Table \ref{tab:extrares}. On Flan-T5-Large, where we do fine-tuning, some  hyperparameters were manually tuned but most left at their default values. The final values for the ones that were manually tuned are provided in Table \ref{tab:training}. 

We perform all experiments with a single NVIDIA Quadro RTX 8000 with 64GB of RAM on an Intel Xeon E502680v4 (2.4GHz).

\begin{table}[]
\centering
\small
\begin{tabular}{@{}lcr@{}}
\toprule
\multicolumn{1}{c}{\textbf{Experiment}}                                                                             & \textbf{Data}  & \multicolumn{1}{c}{\textbf{Cost (US\$)}} \\ \midrule
\multirow{3}{*}{\begin{tabular}[c]{@{}l@{}}Evaluation of \\ Few-Shot In-Context \\ Prompting\end{tabular}} & ADE   & 64.91                           \\
                                                                                                           & CoNLL & 19.24                           \\
                                                                                                           & NYT   & 238.70                          \\ \midrule
\multirow{3}{*}{\begin{tabular}[c]{@{}l@{}}Generation of CoT \\ Explanations (Training Set)\end{tabular}}  & ADE   & 93.96                           \\
                                                                                                           & CoNLL & 44.20                           \\
                                                                                                           & NYT   & 983.86                          \\ \midrule
\begin{tabular}[c]{@{}l@{}}Generation of Target \\ Labels + CoT Explanations\end{tabular}                  & CoNLL & 86.41                           \\ \bottomrule
\end{tabular}
\caption{Summary of costs incurred by prompting and using GPT-3 as a labeler for RE. }
\label{tab:costs}
\end{table}

\subsection{Costs ($\$\$\$$)}
\label{appendix:costs}
We provide details on the costs we incurred while running experiments on GPT-3 in Table \ref{tab:costs}. 

\section{Prompts}
\label{appendix:prompts}
We use the following prompt elements as few-shot exemplars corresponding to each dataset in our evaluation. Inputs and target references are directly extracted from the original training sets while the \texttt{explanations} are human-written and were added when necessary for the experiments described in section \ref{section:gpt3} and \ref{section:flan}.

\subsection*{ADE}
\begin{flushleft}
    \texttt{\textbf{Example Instructional Prefix: } List all [drug, adverse effects] pairs in the TEXT provided below.}
\end{flushleft}
\begin{flushleft}
\texttt{\textbf{TEXT:}  We report on three observations of parkinsonian patients with levo-dopa-induced diphasic dyskinesias, who received subcutaneous apomorphine to reduce the duration of abnormal movements.\\
\textbf{Relations:} [['levo-dopa', 'diphasic dyskinesias']]\\
\textbf{Explanation:} levo-dopa induced diphasic dyskinesias in parkinsonian patients.}<s>
\end{flushleft}
\begin{flushleft}
\texttt{\textbf{TEXT:}  A girl with cystic fibrosis and cyclic neutropenia developed an erythematous papular eruption without fever or neutrophilia 7 months after commencing therapy with G-CSF.\\
\textbf{Relations:} [['G-CSF', 'erythematous papular eruption']]\\
\textbf{Explanation:} G-CSF therapy caused erythematous papular eruption in a girl with cystic fibrosis.<s> }
\end{flushleft}
\begin{flushleft}
\texttt{\textbf{TEXT:}  Hypersensitivity to carboplatin is a rare but real complication of therapy and should be considered in patients presenting with hyperacute changes on ECG whilst receiving carboplatin therapy.\\
\textbf{Relations:} [['carboplatin', 'hyperacute changes on ECG'], ['carboplatin', 'Hypersensitivity']]\\
\textbf{Explanation:} Patients who undergo carboplatin therapy are prone to hypersensitivity and hyperacute changes on their ECG.<s>}
\end{flushleft}
\begin{flushleft}
\texttt{\textbf{TEXT:}  The diagnosis of hypothermia was delayed until it was apparent for several days but resolved with the discontinuation of risperidone and continuation of clozapine.\\
\textbf{Relations: }[['risperidone', 'hypothermia']]\\
\textbf{Explanation:} risperidone caused hypothermia since it was resolved with its discontinuation.<s>}
\end{flushleft}
\begin{flushleft}
\texttt{\textbf{TEXT:}  Eighty-two patients with various malignancies who received imipenem/cilastatin 143 times for neutropenic fever between March 1994 and October 1999 in Department of Pediatric Oncology, Gazi University, were identified.\\
\textbf{Relations:} [['cilastatin', 'neutropenic fever'], ['imipenem', 'neutropenic fever']]\\
\textbf{Explanation:} Patients who received either cilastatin or imipenem were identified with neutropenic fever.<s>}
\end{flushleft}
\begin{flushleft}
\texttt{\textbf{TEXT:}  This increase when clozapine was switched to risperidone and vice versa is consistent with our previous report of elevated serum triglyceride levels in clozapine-treated patients.\\
\textbf{Relations:} [['clozapine', 'elevated serum triglyceride levels']]\\
\textbf{Explanation:} There was a report of elevated serum triglyceride levels in clozapine-treated patients.<s>}
\end{flushleft}
\begin{flushleft}
\texttt{\textbf{TEXT: } Autopsy findings were consistent with bleomycin and oxygen-induced pulmonary damage.\\
\textbf{Relations:} [['bleomycin', 'pulmonary damage'], ['oxygen', 'pulmonary damage']]\\
\textbf{Explanation:} Both bleomycin and oxygen caused pulmonary damage in the autopsy findings.<s>}
\end{flushleft}
\begin{flushleft}
\texttt{\textbf{TEXT:}  CD4 T-lymphocyte depletion, myelosuppression, and subsequent severe infections are the major side effects of fludarabine phosphate therapy.\\
\textbf{Relations: }[['fludarabine phosphate', 'CD4 T-lymphocyte depletion'], ['fludarabine phosphate', 'myelosuppression'], ['fludarabine phosphate', 'severe infections']]\\
\textbf{Explanation: }Following major side-effects are known of fludarabine phosphate therapy, CD4 T-lymphocyte depletion, myelosuppression, and severe infections.<s>}
\end{flushleft}
\begin{flushleft}
\texttt{\textbf{TEXT: } OBJECTIVE: To describe a case of severe skin necrosis resulting from peripheral intravenous administration of low-dose vasopressin in a patient with catecholamine-resistant septic shock.\\
\textbf{Relations:} [['vasopressin', 'skin necrosis']]\\
\textbf{Explanation:} A case of skin necrosis was described after administration of low-dose vasopressin.<s>}
\end{flushleft}
\begin{flushleft}
\texttt{\textbf{TEXT:}  In vitro inhibition of hematopoiesis in a patient with systemic sclerosis treated with D-penicillamine.\\
\textbf{Relations:} [['D-penicillamine', 'inhibition of hematopoiesis']]\\
\textbf{Explanation:} Patient treated with D-penicillamine had in vitro inhibition of hematopoiesis.<s>}
\end{flushleft}
\begin{flushleft}
\texttt{\textbf{TEXT:}  PURPOSE: We report an unusual paradoxical effect of brimonidine.\\
\textbf{Relations:} [['brimonidine', 'paradoxical effect']]\\
\textbf{Explanation:} paradoxical effect of brimonidine was reported.<s>}
\end{flushleft}
\begin{flushleft}
\texttt{\textbf{TEXT:}  Hepatocellular damage following therapeutic intravenous iron sucrose infusion in a child. \\
\textbf{Relations: }[['iron sucrose', 'Hepatocellular damage']] \\
\textbf{Explanation:} Hepatocellular damage occurred in a child after infusion of iron sucrose.<s>}
\end{flushleft}

\subsection*{CoNLL}
\begin{flushleft}
    \texttt{\textbf{Examplee Instructional Prefix: } List the relations of the types [OrgBased In, Work For, Located In, Live In, Kill] among the entities [PERSON, LOCATION, ORGANIZATION, OTHER] in the given text and provide a reasonable explanation.}
\end{flushleft}
\begin{flushleft}
\texttt{\textbf{TEXT:} ``If it does not snow, and a lot, within this month we will have no water to submerge 150,000 hectares (370,500 acres) of rice'', said Bruno Pusterla, a top official of the Italian Agricultural Confederation.\\
\textbf{Relations:}  [['Bruno Pusterla:Per', 'Work For', 'Italian Agricultural Confederation:Org']]\\
\textbf{Explanation:} Bruno Pusterla is a top official of the Italian Agricultral Confederation.<s>}
\end{flushleft}
\begin{flushleft}
\texttt{\textbf{TEXT:} Meanwhile, Shi Liming at the Institute of Zoology of Kunming found that pandas lack variety in their protein heredity, which may serve as one of the major reasons for pandas' near extinction.\\
\textbf{Relations:}  [['Shi Liming:Per', 'Work For', 'Institute of Zoology:Org'], ['Institute of Zoology:Org', 'OrgBased In', 'Kunming:Loc']]\\
\textbf{Explanation:} Shi Liming works for the Institute of Zoology, which is an organization based in Kunming.<s>}
\end{flushleft}
\begin{flushleft}
\texttt{\textbf{TEXT:} The viewers of ``JFK'' and ``The Men Who Killed Kennedy'' never learn about these facts, nor do they ever learn about all of the other massive body of evidence that conclusively proves beyond a reasonable doubt that Oswald was the lone gunman who killed President Kennedy and Officer Tippit and that there was no coverup by Earl Warren or by the Warren Commission.\\
\textbf{Relations:}  [['Oswald:Per', 'Kill', 'President Kennedy:Per'], ['Oswald:Per', 'Kill', 'Officer Tippit:Per']]\\
\textbf{Explanation:} Oswald was the lone gunman who killed President Kennedy and Officer Tippit.<s>}
\end{flushleft}
\begin{flushleft}
\texttt{\textbf{TEXT:} PURCHASE, N.Y .\\
\textbf{Relations:}  [['PURCHASE:Loc', 'Located In', 'N.Y.:Loc']]\\
\textbf{Explanation:} PURCHASE is a place located in N.Y..<s>}
\end{flushleft}
\begin{flushleft}
\texttt{\textbf{TEXT:} BELGRADE, Yugoslavia (AP)\\
\textbf{Relations:}  [['BELGRADE:Loc', 'Located In', 'Yugoslavia:Loc'], ['AP:Org', 'OrgBased In', 'BELGRADE:Loc'], ['AP:Org', 'OrgBased In', 'Yugoslavia:Loc']]\\
\textbf{Explanation:} City of BELGRADE is located in Yugoslavia and AP is an organization based in BELGRADE, Yugoslavia.<s>}
\end{flushleft}
\begin{flushleft}
\texttt{\textbf{TEXT:} Rome is in Lazio province and Naples in Campania. \\
\textbf{Relations:}  [['Rome:Loc', 'Located In', 'Lazio:Loc'], ['Naples:Loc', 'Located In', 'Campania:Loc']]\\
\textbf{Explanation:} Rome is a place located in Lazio and Naples is a place located in Campania.<s>}
\end{flushleft}
\begin{flushleft}
\texttt{\textbf{TEXT:} (By ITAR-TASS correspondent Mikhail Shevtsov)\\
\textbf{Relations:}  [['Mikhail Shevtsov:Per', 'Work For', 'ITAR-TASS:Org']]\\
\textbf{Explanation:} Mikhail Shevtsov is a correspondent for the ITAR-TASS.<s>}
\end{flushleft}
\begin{flushleft}
\texttt{\textbf{TEXT:} In the communique, the Group of Rio states that the Haitian crisis can be resolved only if unrestricted respect is shown for the Governor's Island Agreement which calls for the prompt return of Haitian President Jean Bertrand Aristide to the exercise of his constitutional powers in Haiti. \\
\textbf{Relations:}  [['Jean Bertrand Aristide:Per', 'Live In', 'Haiti:Loc']]\\
\textbf{Explanation: } Jean Bertrand Aristide was the president of Haiti and therefore lived in Haiti.<s>}
\end{flushleft}
\begin{flushleft}
\texttt{TEXT: Moscow ITAR-TASS\\
\textbf{Relations:}  [['ITAR-TASS:Org', 'OrgBased In', 'Moscow:Loc']]\\
\textbf{Explanation: }ITAR-TASS is an organization based in Moscow.<s>}
\end{flushleft}
\begin{flushleft}
\texttt{\textbf{TEXT: }King rose to prominence after Mrs. Parks ' action in December 1955 in Montgomery , Ala. , set the stage for a boycott and subsequent demonstrations that caught the nation by surprise. \\
\textbf{Relations: } [['Mrs. Parks:Per', 'Live In', 'Montgomery:Loc'], ['Mrs. Parks:Per', 'Live In', 'Ala.:Loc'], ['Montgomery:Loc', 'Located In', 'Ala.:Loc']]\\
\textbf{Explanation:} Mrs. Parks actions were in Montgomery, Ala., where she lived. It can be derived that Montgomery is located in Ala..<s>}
\end{flushleft}
\begin{flushleft}
\texttt{\textbf{TEXT:} Sirhan says he was the lone assassin but can't remember shooting Kennedy.\\
\textbf{Relations:}  [['Sirhan:Per', 'Kill', 'Kennedy:Per']]\\
\textbf{Explanation: }Sirhan was the lone assassin in the Kennedy assassination.<s>}
\end{flushleft}
\begin{flushleft}
\texttt{\textbf{TEXT:} In Colorado, 13 inches of snow in Denver Wednesday prompted officials to close Interstate 270 temporarily.\\
\textbf{Relations:}  [['Denver:Loc', 'Located In', 'Colorado:Loc']]\\
\textbf{Explanation:} Denver officials closed Interstate 270 in Colorado, consequently we can see that Denver is located in Colorado.<s>}
\end{flushleft}
\begin{flushleft}
\texttt{\textbf{TEXT:} Edward Marks, an official with the Montgomery County Democratic Party, argued that if Ms. Toth is not interested in the job, ``she should get out''.\\
\textbf{Relations:}  [['Edward Marks:Per', 'Work For', 'Montgomery County Democratic Party:Org']]\\
\textbf{Explanation:} Edward Marks is an official that works for the Montgomery County Democratic Party.<s>}
\end{flushleft}

\subsection*{NYT}
\begin{flushleft}
\texttt{\textbf{TEXT:} Massachusetts ASTON MAGNA Great Barrington; also at Bard College, Annandale-on-Hudson, N.Y., July 1-Aug.\\
\textbf{Relations:}  [['Annandale-on-Hudson', '/location/location/contains', 'Bard College']]\\
\textbf{Explanation:} Annandale-on-Hudson is a location in N.Y. that contains Bard College.<s>}
\end{flushleft}
\begin{flushleft}
\texttt{\textbf{TEXT:} It will be the final movie credited to Debra Hill, a film producer and native of Haddonfield, who produced ``Halloween'' and was considered a pioneering woman in film.\\
\textbf{Relations:}  [['Debra Hill:Per', '/people/person/place-of-birth', 'Haddonfield:Loc']]\\
\textbf{Explanation:} Debra Hill was a film producer born (native of) in Haddonfield.<s>}
\end{flushleft}
\begin{flushleft}
\texttt{\textbf{TEXT:} Under pressure from Mr. Kerkorian and other disgruntled shareholders, Mr. Wagoner started talks on Friday in Detroit with Carlos Ghosn, the chief executive of Renault and Nissan.\\
\textbf{Relations:}  [['Carlos Ghosn:Per', '/business/person/company', 'Renault:Org']]\\
\textbf{Explanation:} Carlos Ghosn is a business person (chief executive) associated with Renault and Nissan.<s>}
\end{flushleft}
\begin{flushleft}
\texttt{\textbf{TEXT:} Mr. Ferrer still holds commanding leads over the other two Democrats in the race -- United States Representative Anthony D. Weiner of Brooklyn and Queens, and City Council Speaker Gifford Miller -- and is also ahead of Mayor Michael R. Bloomberg in most polls.\\
\textbf{Relations:}  [['Anthony D. Weiner:Per', '/people/person/place-lived', 'Brooklyn:Loc'], ['Anthony D. Weiner:Per', '/people/person/place-lived', 'Queens:Loc']]\\
\textbf{Explanation:} Anthony D. Weiner is a person representing Brooklyn and Queens, therefore we can infer he lives in those places.<s>}
\end{flushleft}
\begin{flushleft}
\texttt{\textbf{TEXT:} Quebec, Canada's second most populous province, after Ontario, has not decided to go that far.\\
\textbf{Relations:}  [['Ontario:Loc', '/location/administrative-division/country', 'Canada:Loc'], ['Canada:Loc', '/location/location/contains', 'Ontario:Loc'], ['Canada:Loc', '/location/country/administrative-divisions', 'Ontario:Loc']]\\
\textbf{Explanation:} Ontario is a place located in the administrative divisions of the country Canada. Quebec is Canada's second most populous province and hence, Canada is a place that contains Quebec.<s>}
\end{flushleft}
\begin{flushleft}
\texttt{\textbf{TEXT:} And Abu Izzadeen , who converted to Islam at 17 and heads another successor group to Al Muhajiroun, called Al Ghurabaa, called suicide bombing ``martyrdom operations''.\\
\textbf{Relations:}  [['Abu Izzadeen:Per', '/people/person/religion', 'Islam:Org']]\\
\textbf{Explanation:} Since Abu Izzadeen converted to Islam at the age of 17, we can infer that this is a person who belongs to the religion of Islam.<s>}
\end{flushleft}
\begin{flushleft}
\texttt{\textbf{TEXT:} And yet, despite the success of its exhibitions, the institute remains something of a strange hybrid: located southeast of Notre-Dame, in a striking building designed by Jean Nouvel, it has operated since 1987 as a partnership between France and 22 Arab countries.\\
\textbf{Relations:}  [['Jean Nouvel:Per', '/people/person/nationality', 'France:Loc']]\\
\textbf{Explanation:} Jean Nouvel was a french designer and we can derive his nationality/citizenship as French or France.<s>}
\end{flushleft}
\begin{flushleft}
\texttt{\textbf{TEXT:} They could have done it Sunday, when we were closed,'' said Joseph Bastianich, who owns Del Posto with his mother, Lidia Bastianich, and the chef, Mario Batali.\\
Relations:  [['Lidia Bastianich:Per', '/people/person/children', 'Joseph Bastianich:Per']]\\
Explanation: Joseph Bastianich owns Del Posto with his mother Lidia Bastianich.<s>}
\end{flushleft}
\begin{flushleft}
\texttt{\textbf{TEXT:} A French court sentenced six Algerian-French men to prison terms of up to 10 years on Tuesday for their role in a 2001 plot to attack the United States Embassy in Paris , closing the books on one of France 's most serious terrorist cases.\\
\textbf{Relations: } [['Paris:Loc', '/location/ administrative-division/country', 'France:Loc'], ['France:Loc', '/location/location/contains', 'Paris:Loc'], ['France:Loc', '/location/country/ administrative-divisions', 'Paris:Loc'], ['France:Loc', '/location/country/capital', 'Paris:Loc']]\\
\textbf{Explanation:} Paris is located in the administrative divisons of the country France. Consequently, France is a place that contains Paris. US embassies are located in the capital of countries, therefore it can be inferred that Paris is the capital of France.<s>}
\end{flushleft}
\begin{flushleft}
\texttt{\textbf{TEXT:} Anheuser-Busch, which has been the exclusive beer sponsor for the Super Bowl since 1989, will do so again for the Super Bowls in 2007 and 2010 on CBS and in 2008 and 2011 on Fox Broadcasting , said Anthony T. Ponturo, vice president for global media and sports marketing at Anheuser-Busch in St. Louis.\\
\textbf{Relations:}  [['Anheuser-Busch:Org', '/business/company/place-founded', 'St. Louis:Loc'], ['St. Louis:Loc', '/location/location/contains', 'Anheuser-Busch:Org']]\\
\textbf{Explanation:} Anheuser-Busch is a business that was founded in St. Louis. Consequently, St. Louis is a place that contains Anheuser-Busch.<s>}
\end{flushleft}
\begin{flushleft}
\texttt{\textbf{TEXT:} Somewhat chastened by his retreat in the polls, Mr. Blair acknowledged that Britons had turned against him in part over accusations that he led them into a war in Iraq on dubious legal grounds and on the false premise that Saddam Hussein presented a direct threat because of a supposed arsenal of unconventional weapons that was never found.''\\
Relations:  [['Saddam Hussein:Per', '/people/deceased-person/place-of-death', 'Iraq:Loc'], ['Saddam Hussein:Per', '/people/person/place-of-birth', 'Iraq:Loc'], ['Saddam Hussein:Per', '/people/person/nationality', 'Iraq:Loc']]\\
\textbf{Explanation:} Saddam Hussein was killed in Iraq. His place of birth was also Iraq. We can infer that his nationality was Iraq.<s>}
\end{flushleft}
\begin{flushleft}
\texttt{\textbf{TEXT:} Rupert Murdoch and John C. Malone , who have wrangled for two years over Mr. Malone 's challenge to Mr. Murdoch 's control of the News Corporation , have made peace .
Relations:  [['Rupert Murdoch', '/business/person/company', 'News Corporation'], ['News Corporation', '/business/company/founders', 'Rupert Murdoch']]
Explanation: Rupert Murdoch is a business person associated with News Corporation, which was a company founded by Rupert Murdoch.<s>}
\end{flushleft}
\begin{flushleft}
\texttt{\textbf{TEXT:} Manhattan, especially the East Village , has long been well stocked with cheap and raucous yakitori places that specialize in skewers and beer.\\
\textbf{Relations:}  [['Manhattan:Loc', '/location/location/contains', 'East Village:Loc'], ['East Village:Loc', '/location/neighborhood/neighborhood-of', 'Manhattan:Loc']]\\
\textbf{Explanation:} East Village is a neighborhood in Manhattan.<s>}
\end{flushleft}
\begin{flushleft}
\texttt{\textbf{TEXT:} HEADING OUT -- Sanford I. Weill stepped down as chairman of Citigroup , the worldwide financial supermarket he had meticulously and single-mindedly stitched together through dozens of mergers and acquisitions.\\
\textbf{Relations:}  [['Citigroup:Org', '/business/company/advisors', 'Sanford I. Weill:Per']]\\
\textbf{Explanation: }Citigroup is a business company who was associated with (advised by) Sanford I. Weill.<s>}
\end{flushleft}
\begin{flushleft}
\texttt{\textbf{TEXT:} He had decided to use the premiere to publicize the issue; his plan was to invite the neighborhood's Russian speakers to sign a petition against piracy, a common practice at the area's Russian-language video outlets, which sell films and music from Russia and by Russian immigrants in the United States.\\
\textbf{Relations: } [['Russian:Per', '/people/ethnicity/ geographic-distribution', 'Russia:Loc']]\\
\textbf{Explanation:} Russian is an ethnicity in United States associated with immigrants who came from the geographic distribution of Russia.<s>}
\end{flushleft}
\begin{flushleft}
\texttt{\textbf{TEXT:} In 1995, Cleveland successfully lobbied to have the name Cleveland Browns stay in that city after that venerable franchise's owner, Art Modell, opted to move it to Baltimore.\\
\textbf{Relations:}  [['Cleveland:Loc', '/sports/sports-team-location/teams', 'Cleveland Browns:Org'], ['Cleveland Browns:Org', '/sports/sports-team/location', 'Cleveland:Loc']]\\
\textbf{Explanation:} Cleveland Browns is the sports franchise located in Cleveland, consequently Cleveland's sports team is Cleveland Browns.<s>}
\end{flushleft}
\begin{flushleft}
\texttt{\textbf{TEXT:} Mr. Fields, speaking from vacation in France, added, ``That a mogul like Sumner Redstone could make a statement so vicious, so pompous, so petulant as that he didn't want to make a deal with Tom Cruise because of his personal conduct -- it tells you more about Sumner Redstone and Viacom, than about Tom Cruise''.\\
\textbf{Relations:}  [['Sumner Redstone:Per', '/business/ company-shareholder/ major-shareholder-of', 'Viacom:Org']]\\
\textbf{Explanation:} Sumner Redstone is a major shareholder of the company Viacom.<s>}
\end{flushleft}
\begin{flushleft}
\texttt{\textbf{TEXT:} It is a room of paintings by Leonard Peltier , a citizen of the Anishinabe and Dakota and Lakota nations who is serving two consecutive life terms in Pennsylvania for the murder of two F.B.I. agents on the Pine Ridge Reservation in South Dakota.\\
\textbf{Relations:}  [['Leonard Peltier:Per', '/people/person/ethnicity', 'Lakota:Per'], ['Lakota:Per', '/people/ethnicity/people', 'Leonard Peltier:Per']]\\
\textbf{Explanation:} Leonard Peltier is a member of the Lakota native-american tribe and consequently belongs to that ethnic group.<s>}
\end{flushleft}
\begin{flushleft}
\texttt{\textbf{TEXT:} INSIDE THE N.B.A. Correction : February 9 , 2006 , Thursday A sports article on the Spotlight page on Sunday about Dick Bavetta , a longtime referee in the National Basketball Association, misstated the number he was approaching to set the record for regular-season games worked.\\
\textbf{Relations:}  [['Dick Bavetta:Per', '/people/person/profession', 'National Basketball Association:Org']]\\
\textbf{Explanation: }Dick Bavetta is a person who's profession is that of a referee in National Basketball Association.<s>}
\end{flushleft}

\begin{flushleft}
\texttt{\textbf{TEXT:} Now the United States Postal Service may be displaying a similar rebellious streak : tomorrow at the huge Sturgis motorcycle rally in the Black Hills of South Dakota, the Postal Service will issue a set of four stamps that depict classic American bikes.\\
\textbf{Relations:}  [['United States Postal Service:Org', '/business/company/industry', 'Postal Service:Org']]\\
\textbf{Explanation:} United States Postal Service is a business company in the industry of providing postal services.<s>}
\end{flushleft}

\begin{figure}[h]
    \centering
\includegraphics[scale=0.06]{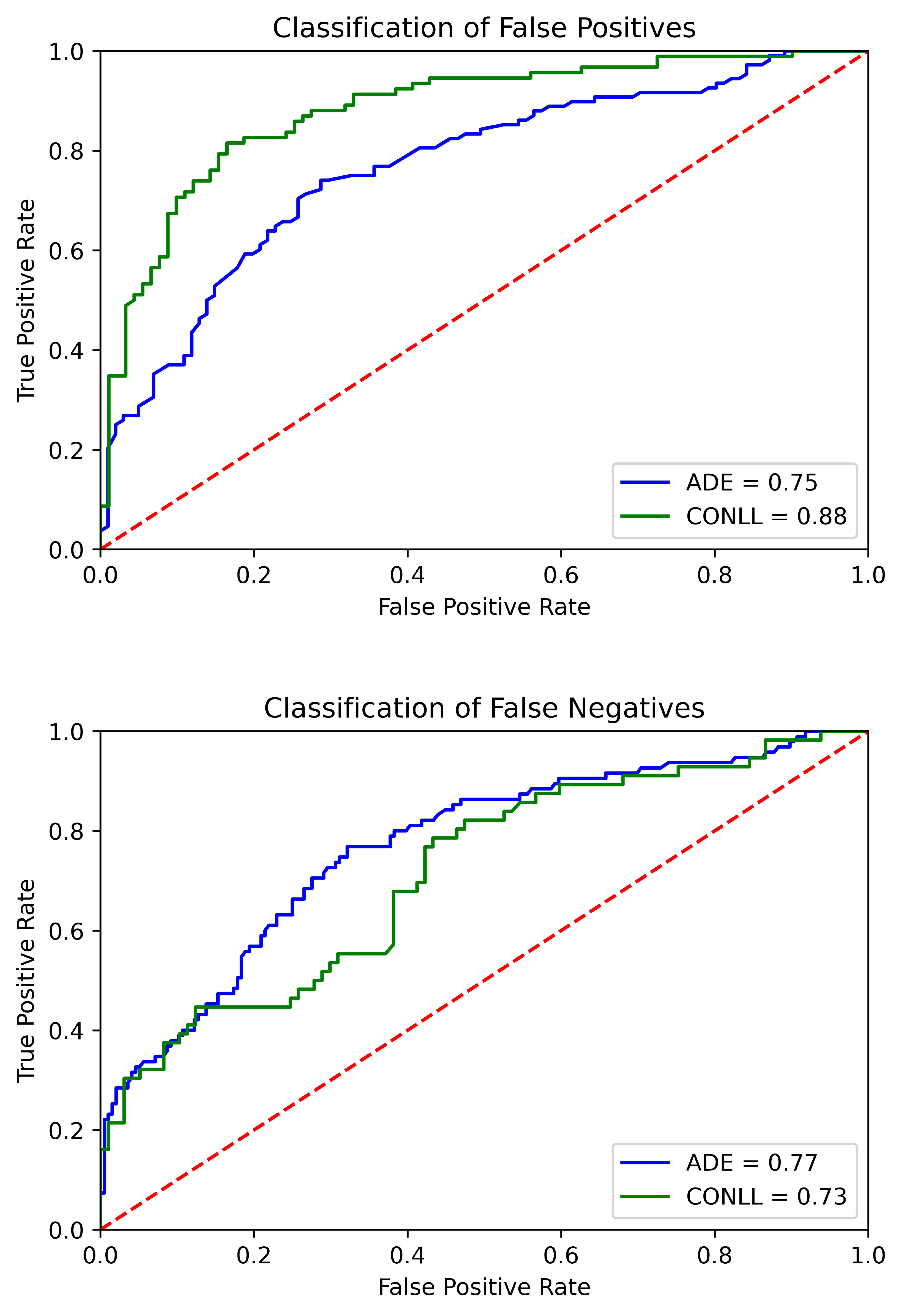}
\caption{AUC plots for FPs and FNs. }
\label{fig:aucroc}
\end{figure}

\begin{table}[]
\centering
\begin{tabular}{@{}lrr@{}}
\toprule
\multicolumn{1}{c}{\textbf{Data}} & \begin{tabular}[c]{@{}r@{}}\textbf{Inaccure FPs}\\ \textbf{/ Total FPs}\end{tabular} & \begin{tabular}[c]{@{}r@{}}\textbf{Inaccurate FNs}\\\textbf{ / Total FNs}\end{tabular} \\ \midrule
ADE                      & 108 / 209                                                          & 136 / 417                                                            \\
CoNLL                    & 92 / 183                                                           & 56 / 152                                                             \\ \bottomrule
\end{tabular}
\caption{Number of inaccurate false positives (FNs) and false negatives (FNs) identified during automated evaluation in GPT-3 labelled outputs under the in-context few-shot prompting setting. }
\label{tab:fpfnfindings}
\end{table}

\section{Learning to Identify \emph{False} False Positives and Negatives}
\label{appendix:evaluations}
As discussed in the main paper, one common problem across datasets in generative RE is evaluation, given that LMs are flexible in how they might express entities and relations. 
Prior work in RE has tended rely on standard metrics to quantify performance (precision, recall, micro-F1). These rely on matching \textit{classified} (or in our case, \textit{generated}) labels to reference labels to calculate the number of true positives (TPs), false positives (FPs), true negatives (TNs), and false negatives (FNs). 

Prior to the introduction of LLMs for generative RE, \citet{taille-etal-2020-lets} attempted to unify evaluation and provide useful guidelines around issues associated with prior methods and how different evaluation strategies rendered an accurate comparison infeasible. They broadly recommended the use of a \textit{strict} evaluation scheme where for a relation triplet to be considered correct, the head and tail entity surface forms must be an exact match, as well as their corresponding types (when available). While this  provides a standardized framework for traditional models where entities and and relations are hard \textit{classification} labels, in a generative setting we often find that LLMs, under varying levels of supervision, produce relation triplets (or pairs) that do not correspond exactly to their reference counterparts, but are nonetheless correct upon manual review. 
Consider the following example from CoNLL in Figure \ref{fig:examples}
\begin{flushleft}
\texttt{\textbf{Text: }On Friday, U.S. Ambassador Vernon A. Walters... fuselage.}

\texttt{\textbf{Gold Reference:} [(Vernon A. Walters, `Live In', U.S.)]}

\texttt{\textbf{\textit{Generated }Relations: }[[Vernon A. Walters, `Works For', U.S.]]}
\end{flushleft}

\vspace{0.5em}
\noindent In this example, one can reasonably infer that \texttt{Vernon A. Walter} is a U.S. Ambassador. Therefore, by definition a U.S. diplomat to another country cannot live inside the U.S., but such a person must work for the U.S. (commonsense dictates that a diplomat would work for a specific country). 

To achieve a more accurate characterization of how LLMs perform on generative RE tasks, we hired human annotators on Amazon Mechanical Turk\footnote{We set the payrate to average at $\$15$/hour using time estimates informed by pilot experiments.} to manually re-assess all ostensible FPs and FNs from each of our datasets. To control for quality and recruit annotators we ran pilot experiments on 50 instances of pre-annotated data.\footnote{These instances used in pilot experiments were annotated by a graduate student familiar with this research.} We required AMT workers to have an overall approval rating of $>95\%$ irrespective of geographic region. Based on these initial set of results we hired a total of 9 workers who reliably followed our instructions. Recruited workers were paid periodic bonuses (equivalent to one hour of pay) based on the quality of their annotations.

To identify potentially faulty ``false positives'', we provided annotators with the input text along with the relation identified as a FP, and ask the following question: ``Can the given given relation be reasonably derived from the text?''. 
Similarly, to identify erroneous ``false negatives'', we provide annotators with the input text, the full set of generated labels, the \textit{ostensible} FN from the reference set, and ask: ``Can the reference relation triplet (or pair) be inferred from the generated set of relations?''. 
Each instance was annotated by three different AMT workers, and we considered a potential FP/FN to be inaccurate only when \textbf{all} annotators agree on a label.\footnote{We observe a high degree of agreement among the annotators with a Fliess $kappa$ of 0.83} 
We provide specific examples of FPs and FNs in Tables \ref{tab:fp} and \ref{tab:fn}. We summarize the dataset-specific findings in Table \ref{tab:fpfnfindings}. 

In light of these findings, we make a first effort in using simple, learned models to classify false-positives/negatives in generative RE. We experiment with fine-tuned BERT \cite{devlin-etal-2019-bert} classifier to classify ``false positives'' and ``false negatives'' as being accurate designations (or not). 
For FPs, we concatenate the input with a generated relation pair/triplet (\textit{potential} FP) and classify using the \texttt{[CLS]} token -
\begin{flushleft}
\centering
    \texttt{[CLS] Input Text [SEP] Potential FP}
\end{flushleft}
Similarly, for FNs we concatenate the input text with a \textit{potential FN} and the full set of generated labels, and classify using the \texttt{[CLS]} token - 
\begin{flushleft}
\centering
    \texttt{[CLS] Input Text [SEP] Potential FN [SEP] Generated Labels}
\end{flushleft}

We analyze the effectiveness of this approach in Figure \ref{fig:aucroc} using the AUC-ROC. We find that this approach is most effectiveness in identifying potential potential false positives for CoNLL (AUC $0.88$), while being least effective at identifying false negatives for CoNLL (AUC $0.73$). 
This suggests that learning to identify erroneous ``false positives'' and ``false negatives'' may be a promising avenue to facilitate accurate automated evaluation of generative LLMs for RE.

\begin{table*}[]
\renewcommand*{\arraystretch}{1.5}
\large
\centering
\begin{adjustbox}{width=1.5\textwidth, angle=90}
\begin{tabular}{@{}cp{6cm}cc@{}}
\hline
\textbf{Dataset}                  & \multicolumn{1}{c}{\textbf{Input}}                                                                                                                                                                & \begin{tabular}[c]{@{}c@{}}\textbf{Detected FN}\\ (From the Gold Reference Set)\end{tabular}                                                                                         & \begin{tabular}[c]{@{}c@{}}\textbf{Full Set of} \\ \textbf{Generated Relations}\end{tabular}                                                \\ \hline
\multirow{6}{*}{CoNLL04} & $\bullet$ The three reactors, all at the Savannah River Plant, in Aiken, S.C., have been shut down since last April undergoing changes to make them safer.                                         & {[}Savannah River Plant, Located\_In, S.C.{]}                                                                                                                     & \begin{tabular}[c]{@{}c@{}}{[}Savannah River Plant, Located\_In, Aiken{]} \\ {[}Aiken, Located\_In, S.C.{]}\end{tabular}      \\
                         & $\bullet$ They will also be cleaning the car Oswald drove on the day Kennedy was shot and the ambulance that took Oswald to hospital after he was shot by Jack Ruby.                               & {[}'Jack Ruby', '\textbf{Kill}', 'Oswald'{]}                                                                                                                               & {[}Jack Ruby, \textbf{Shoot}, Oswald{]}                                                                                              \\ \hline
\multirow{7}{*}{ADE}     & $\bullet$ We report the case of an 11-year-old female treated for mediastinal t-cell lymphoma who presented renal failure following the second cycle of high-dose methotrexate (hdmtx).            & {[}'hdmtx', 'renal failure'{]}                                                                                                                                    & {[}'methotrexate', 'renal failure'{]}                                                                                       \\
                         & $\bullet$ Four days after the initial injection of 3.6 mg of goserelin acetate, severe dyspnea developed due to worsening pleuritis carcinomatosa, which was considered as a flare-up.                                                                           & {[}'goserelin acetate', 'flare'{]}                                                                                                                                      & {[}'goserelin acetate', 'dyspnea'{]}                                                                               \\ \hline
\multirow{8}{*}{NYT}     & $\bullet$ This time, the president chose Father Leon to replace one of those clergymen, the Rev. Franklin Graham, who was filling in for his father, the Rev. Billy Graham, who was ill in 2001.   &  {{[}'Franklin Graham', '\textbf{people/people/children}', 'Billy Graham'{]}}                                                                                                       & \multicolumn{1}{l}{{[}'Billy Graham', '\textbf{/people/person/children}', 'Franklin Graham'{]}  }                                                    \\
                         & $\bullet$ The moves by Citigroup and Deutsche Bank are only the latest illustrations of the difficulty of retaining jobs in New York City and rebuilding the business district in Lower Manhattan. & \begin{tabular}[c]{@{}c@{}}{[}'Citigroup', 'business/place/founded', 'New York City'{]}\\ {[}'Lower Manhattan', '\textbf{/location/location/contains}', 'New York City'{]}\end{tabular} & {[}'New York City', '\textbf{/location/location/contains}', 'Lower Manhattan'{]}                                                                   \\ \hline
\end{tabular}
\end{adjustbox}
\caption{Sample of inaccurate false \textit{negatives} identified by human annotators for each dataset. Examples from CoNLL and ADE were generated from GPT-3, while those in NYT were generated by Flan.}
\label{tab:fn}
\end{table*}

\begin{table*}[]
\centering
\begin{adjustbox}{width=1.5\textwidth, angle=90}
\small
\begin{tabular}{@{}cp{6cm}cc@{}}
\toprule
Dataset                  & \multicolumn{1}{c}{Input}                                                                                                                                                                                      & \begin{tabular}[c]{@{}c@{}}Detected FPs \\ (Generated)\end{tabular}                                                                                                                & \begin{tabular}[c]{@{}c@{}}Full Set of \\ True Relations\end{tabular}                                        \\ \midrule
\multirow{14}{*}{CoNLL04} & $\bullet$ Illinois Gov. James Thompson signed legislation last month scrapping Chicago 's central school board next July , to be replaced by parent-run councils empowered to set budgets and hire and fire principals . & \begin{tabular}[c]{@{}c@{}}{[}James Thompson, \textbf{Work\_For}, Illinois{]},\\  {[}Chicago, \textbf{Located\_In}, Illinois{]}\end{tabular} & {[}{[}James Thompson, \textbf{Live\_In}, Illinois{]}{]}                  \\
                         & $\bullet$ On Friday, U.S. Ambassador Vernon A. Walters displayed photographs of one Libyan jet showing shapes resembling missile pods on its wings and fuselage.                                                       & {[}Vernon A. Walters, \textbf{Work\_For}, U.S.{]}                                                                                         & {[}{[}Vernon A. Walters, \textbf{Live\_In}, U.S.{]}                         \\ \midrule
\multirow{10}{*}{ADE}     & $\bullet$ Background: how to best treat psychotic patients who have had past clozapine-induced agranulocytosis or granulocytopenia remains a problem.                                                                    & {[}clozapine, agranulocytosis{]}                                                                                                   & {[}{[}clozapine, granulocytopenia{]}{]}                             \\
                         & $\bullet$ Acute renal failure is a rare complication following the administration of intravenous immunoglobulin (ivig).                                                                                                  & {[}ivig, acute renal failure{]}                                                                                                    & {[}{[}immunoglobulin, acute renal failure{]}{]}                      \\ \midrule
\multirow{11}{*}{NYT}     & $\bullet$ Many people in Turkey have lost hopes in joining Europe and they are looking for other horizons , '' said Onur Oymen , an opposition politician whose party is staunchly secular.                              & {[}Turkey, \textbf{location/administrative\_division/country}, Europe{]}                                                                                               & \multicolumn{1}{l}{{[}{[}Europe, \textbf{LOC\_CONTAINS}, Turkey{]}{]}}      \\
                         & $\bullet$ To make his case , Dr. von Hagens invited two journalists to Dalian for a tour of his facility , which he said was the first center in China to preserve bodies.                                               & {[}Dalian, \textbf{location/administrative\_division/country}, China{]}                                                                                                & {[}{[}China, \textbf{location/location/contains}, Dalian{]}{]}                           \\ \bottomrule
\end{tabular}
\end{adjustbox}
\caption{Sample of inaccurate false \textit{positives} (FPs) identified by human annotators for each dataset. Examples from CoNLL and ADE were generated from GPT-3, while those in NYT were generated by Flan.}
\label{tab:fp}
\end{table*}

\subsection{List of out-of-domain relation-types generated by Flan during Few-Shot Prompting with CoNLL}
\label{appendix:evaluationsconlllist}
Assassinates, Purpose, isPartOf, Mother, Spouse, President, date, killed, Summer, Works\_at, Sentenced\_To\_Death, Source, Statue, Secretary, Born, Year, Born\_in, Day, Place, Number\_Of\_Passengers, Callers, Governor, Hometown, has\_a\_leader, is\_a\_member\_of, Nickname, is\_part\_of, Office, Rank, Works\_For, WorkedFor, Worked\_For, Killed\_By, Piano, Term, Sentence, Person, Movie, Said, Brother, Date\_of\_Death, Type, Death\_Penalty, assassination\_date, Worked\_for, capital, Killed, Killing, Occupation, Crime, Years\_in\_use, Org, Education, Order\_to\_ignore, Assassination, Location, Officer, language, former\_name, Total\_acres, Age, Cause, Chairman, worked\_for, Son, Staff\_name, departure, Capsule\_name, Operator, Spin-off, Owner, located\_in, theory, Birth\_Place, on\_duty\_with, City, Top\_Leader, Director, structure, Known\_as, former\_chief\_executive, Works\_for, Native\_name, Percentage, department, Component, reminds\_someone\_of, Sex, Bank, Appointed\_By, Activity, Title, has\_a\_river\_name, Size, Office\_Space, Part, Kingdom, Attached\_to, Death\_Place, Years\_on\_the\_Supreme\_Court, Assassin, location, Newspaper, City,, island, Employee, Friend, Native\_Son, Speaker, Visitor, Date, Aircraft, channel, Sale\_to, Creditor, Client, Nationality, Flight\_Status, assassinater, on\_behalf\_of, Shot\_By.

\end{document}